\documentclass[journal]{IEEEtran}
\usepackage{amsmath,amsfonts}
\usepackage{algorithm}
\usepackage{array}
\usepackage[caption=false,font=scriptsize,labelfont=sf,textfont=sf]{subfig}
\usepackage{textcomp}
\usepackage{stfloats}
\usepackage{url}
\usepackage{verbatim}
\usepackage{graphicx}
\usepackage{cite}
\usepackage{bm}
\usepackage{multirow}
\usepackage{threeparttable}
\usepackage{booktabs}
\usepackage{wrapfig}
\usepackage{tabularx}
\usepackage{algpseudocode}

\usepackage{etoolbox}
\makeatletter
\patchcmd{\@makecaption}
{\scshape}
{}
{}
{}
\makeatother

\begin{document}

\title{CR-LSO: Convex Neural Architecture Optimization in the Latent Space of Graph Variational Autoencoder with Input Convex Neural Networks}

\author{Xuan Rao, Bo Zhao, \IEEEmembership{Senior Member, IEEE}, and Derong Liu, \IEEEmembership{Fellow, IEEE}
	\thanks{\emph{Corresponding author: Bo Zhao}}
	\thanks{This work was supported in part by the National Natural Science Foundation of China under Grant 61973330, in part by the Open Research Project of the Key Laboratory of Industrial Internet of Things and Networked Control, Ministry of Education under Grant 2021FF10, in part by the Fundamental Research Funds for the Central Universities under grant 1243300008, and in part by the Beijing Normal University Tang Scholar.}
	\thanks{Xuan Rao and Bo Zhao are with the School of Systems Science, Beijing Normal University, Beijing 100875, China (e-mail: raoxuan98@mail.bnu.edu.cn; zhaobo@bnu.edu.cn).}
	\thanks{Derong Liu is with the School of System Design and Intelligent Manufacturing, Southern University of Science and Technology, Shenzhen 518000, China (email: liudr@sustech.edu.cn), and also with the Department of Electrical and Computer Engineering, University of Illinois Chicago, Chicago, IL 60607, USA (e-mail: derong@uic.edu).}
	
	\thanks{© 2025 IEEE. Personal use of this manuscript is permitted. Permission from IEEE must be obtained for all other uses, in any current or future media, including reprinting/republishing this material for advertising or promotional purposes, creating new collective works, for resale or redistribution to servers or lists, or reuse of any copyrighted component of this work in other works.}
}

\markboth{IEEE Transaction on Emerging Topics in Computational Intelligence}%
{}


\maketitle
\newtheorem{remark}{Remark}

\begin{abstract}
In neural architecture search (NAS) methods based on latent space optimization (LSO), a deep generative model is trained to embed discrete neural architectures into a continuous latent space. In this case, different optimization algorithms that operate in the continuous space can be implemented to search neural architectures. However, the optimization of latent variables is challenging for gradient-based LSO since the mapping from the latent space to the architecture performance is generally non-convex. To tackle this problem, the present paper develops a convexity regularized latent space optimization (CR-LSO) method, which aims to regularize the learning process of latent space in order to obtain a convex architecture performance mapping. Specifically, CR-LSO trains a graph variational autoencoder (G-VAE) to learn the continuous representations of discrete architectures. Simultaneously, the learning process of latent space is regularized by the guaranteed convexity of input convex neural networks (ICNNs). In this way, the G-VAE is forced to learn a convex mapping from the architecture representation to the architecture performance. Hereafter, the CR-LSO approximates the performance mapping using the ICNN and leverages the estimated gradient to optimize neural architecture representations. Experimental results on three popular NAS benchmarks show that CR-LSO achieves competitive evaluation results in terms of both computational complexity and architecture performance\footnote{Codes are available at https://github.com/RaoXuan-1998/CR-LSO.}.
\end{abstract}
\begin{IEEEkeywords}
Neural architecture search, latent space optimization, variational auto-encoder, input convex neural network, graph-based optimization.
\end{IEEEkeywords}

\section{Introduction}
\label{sec:intro}
Many real-world problems, such as the function network design \cite{DBLP:conf/nips/AstudilloF21}, the functional protein prediction (e.g., the AlphaFOLD \cite{jumper2021highly}), etc., can be formulated as optimizing graphs that represent proper computational performance on given tasks. Neural architecture search (NAS) can also be described as a graph optimization problem since a neural architecture is a directed acyclic graph (DAG) from the view of data streams. Thus, finding an optimal architecture is equivalent to determining the optimal topology of the DAG. However, many efficient optimization methods, such as Bayesian optimization (BO), simulated annealing, and gradient-based optimization, which operate in continuous space primarily, are not applicable to the optimization of neural architectures directly due to the discreteness of graphs.

To optimize neural architectures in the continuous space, this paper resorts to a special technique called latent space optimization (LSO), where a deep generative model is trained to embed the discrete architecture into a relatively smooth latent space. In this way, the search space of NAS is transformed from discrete to continuous. The objective of LSO-based NAS can be formulated as  
\begin{align}
	\label{eq:lso}
	\begin{split}
		& {\rm max}_{z\in \mathcal{Z}} \ \ {\rm Performance} (\alpha) \\
		& {\rm s.t.} \ \ \ \alpha = {\rm Decode}(z) \ \ {\rm and} \ \ \alpha \in \mathcal{I},
	\end{split}
\end{align}
where $\mathcal{I}$ is the discrete search space, $\mathcal{Z}$ is the continuous latent space, and ${\rm Decode}(z)$ is the decoding function which converts the continuous representation $z$ to the discrete architecture $\alpha$. From \eqref{eq:lso}, we know that the architecture is searched in the continuous space $\mathcal{Z}$ by optimizing $z$ but is evaluated in the discrete space $\mathcal{I}$ by evaluating $\alpha$. Although some applications of LSO have received success in NAS tasks \cite{luo2018neural, DBLP:conf/nips/ZhangJCGC19, chatzianastasis2021graph}, most of them discussed how to design an encoder-decoder model to better learn the continuous representations of neural architectures, and only a few of them discussed the subsequent optimization method for architecture latent variables. 

This paper focuses on improving the effectiveness of gradient-based LSO for architecture search \cite{luo2018neural}, where the gradient of the architecture performance with respect to the latent variable is estimated by a performance predictor. However, gradient-based optimization is challenging as described below. First, the mapping from the latent space to the architecture performance is generally non-convex. Although the similarity of different architectures can be measured directly in the Euclidean latent space of deep generative models, the performance mapping is not guaranteed to be a convex function, which raises challenges for the gradient-based optimization. Second, most conventional function approximators, e.g., the multi-layer perceptron (MLP), the graph neural network (GNN), the radial basis function (RBF), etc., are not convex from input to output. It means that even if the underlying performance mapping is convex, one may not obtain a convex predictor due to the non-convexity of function approximators. 

\begin{figure*}[htbp]
	\centering
	\includegraphics[width=0.85\textwidth]{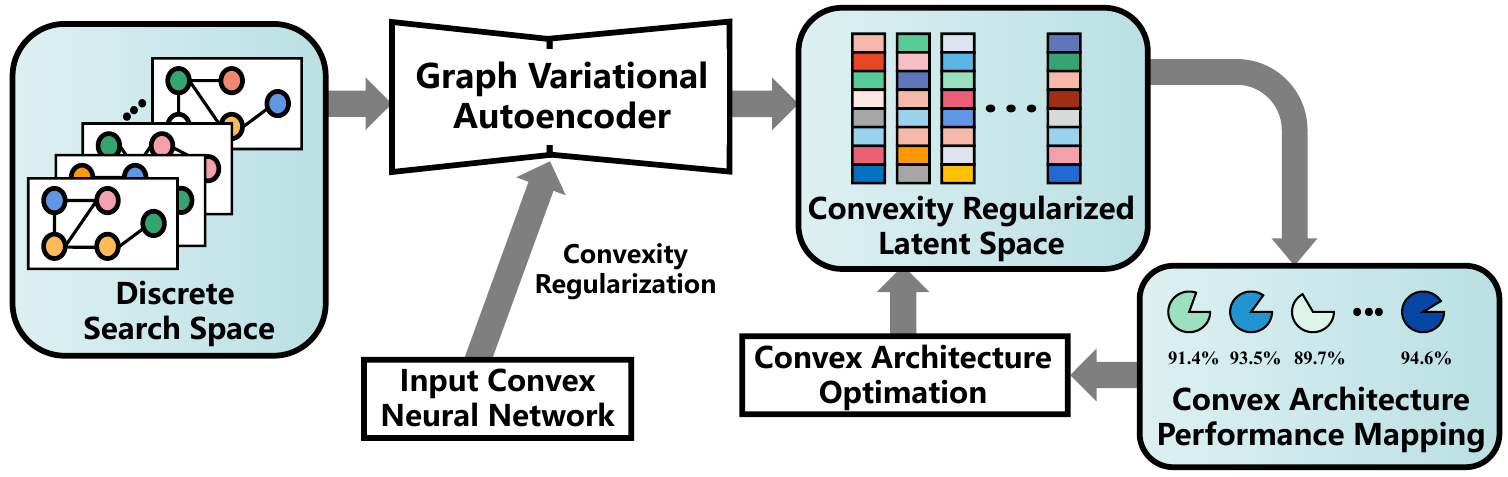}
	\caption{The conceptual visualization of the proposed CR-LSO. By using a graph variational autoencoder, CR-LSO transforms the discrete search space of NAS into a continuous latent space. Simultaneously, CR-LSO utilizes the guaranteed convexity of the ICNN to regularize the learning process of latent space, so as to obtain a convex architecture performance mapping, which makes the gradient-based LSO more effective.}
	\label{fig:cr_lso}
\end{figure*}

Can the (approximate) optimality of gradient-based LSO be guaranteed? This paper tries to answer this question by developing a convexity regularized latent space optimization (CR-LSO) method, whose main process is shown in Fig. \ref{fig:cr_lso}. First, the CR-LSO builds a graph variational autoencoder (G-VAE) to learn the continuous representations of discrete architectures. Simultaneously, the learning process of latent space is regularized by the guaranteed convexity of input convex neural networks (ICNNs) \cite{amos2017input}. In this way, the G-VAE has to learn a (approximate) convex performance mapping. Otherwise, the prediction loss of architecture performance cannot decrease to a sufficiently low level.\footnote{Note that ICNNs are capable of approximating only convex functions. Consequently, if the mapping representing architecture performance is non-convex, the prediction error will be substantial. To address this, if the goal is to minimize the prediction loss, the embedding methodology employed in G-VAE must be adapted in such a way as to ensure that the induced mapping is rendered convex. This implies implementing a form of regularization in learning the latent space by leveraging the guaranteed convexity of ICNNs.} However, the learning process of convexity regularized latent space requires amounts of labeled architectures. To avoid this problem, a GNN-based predictor is built to predict the performance of unlabeled architectures. Thus, the learning process of latent space is indeed based on a semi-supervised manner. Hereafter, the gradient optimization of latent variables is executed with a convex performance predictor in a convex space. The main contributions of this paper are summarized as follows.
\begin{itemize}
	\item Neural architectures are searched in the continuous space by learning a G-VAE which embeds the discrete architecture into the latent space. In this case, the similarity of neural architectures can be utilized to improve the search efficiency of NAS methods.
	\item The CR-LSO method, a convex architecture optimization framework, is developed to improve LSO-based NAS. By regularizing the learning process of the latent space of G-VAE using the guaranteed convexity of the ICNN, the architecture performance mapping is formalized as a convex function, which makes the gradient-based LSO more competitive.
	\item Experimental results on three NAS benchmarks, i.e., the NAS-Bench-101 \cite{DBLP:conf/icml/YingKCR0H19}, the NAS-Bench-201 \cite{DBLP:conf/iclr/Dong020}, and the NAS-Bench-301 \cite{siems2020bench}, show that the proposed CR-LSO achieves competitive NAS results compared to other state-of-the-art NAS methods and hyper-parameter optimization algorithms in terms of computational complexity and architecture performance.
\end{itemize}

\section{Related Works}
\subsection{Input convex neural network (ICNN)}
An ICNN is a scalar-valued neural network with constraints on network parameters such that the network mapping from input to output is a convex function. Originally, it was used to allow the convex inference in various problems, such as structured prediction, data imputation, and continuous action reinforcement learning \cite{amos2017input}. Furthermore, a convex optimal control for complex systems is achieved with implementing ICNNs \cite{chen2018optimal}, and the optimal transport between two distributions is improved by ICNNs \cite{DBLP:conf/icml/MakkuvaTOL20}. 
\subsection{Neural architecture search (NAS)}
Deep neural networks (DNNs) have made substantial breakthroughs in many fields by capturing features automatically. It has transformed the research paradigm in deep learning towards the design of neural architectures. NAS aims to advance the development of DNNs by automatically designing architectures that facilitate the performance on given tasks \cite{he2021automl, elsken2019neural, DBLP:journals/tetci/LinFCTL22, 10005101}. Researches in NAS can be broadly categorized into three main aspects, i.e., the search space, optimization strategies, and the evaluation process.

Typically, an architecture can be represented by a DAG. However, the DAG structure results in a discrete search space, which in turn constrains the direct application of gradient-based optimization techniques. This characteristic prompted early NAS methods to explore black-box optimization approaches, such as reinforcement learning (RL) \cite{zoph2016neural, DBLP:journals/ivc/JaafraLDN19, DBLP:conf/icml/PhamGZLD18}, EA \cite{real2019regularized, DBLP:conf/eccv/NingZZWY20, liu2021survey}, and BO \cite{zhou2019bayesnas, DBLP:conf/aaai/WhiteNS21, DBLP:conf/iclr/RuW0O21}. 
For instance, RL-driven NAS methods may use recurrent neural networks (RNNs) as controllers to generate strings that encode unique neural architectures, evaluate their performance, and refine the controllers based on the feedback signals. 

Later, differentiable architecture search (DARTS) is proposed to improve the NAS efficiency \cite{liu2018darts, xu2021partially, chen2021progressive, DBLP:conf/bmvc/Chu021}. The original DARTS formulates the NAS as a bi-level optimization problem and relaxes the discrete search space by replacing the operator selection with the soft-mixture of all candidate operators to realize gradient-based NAS \cite{liu2018darts}. Subsequent works focused on improving the DARTS efficiency, such as P-DARTS \cite{ chen2021progressive}, which  introduces a path dropout strategy to enhance the robustness of the searched architectures, and PC-DARTS \cite{xu2021partially}, which proposes a partial channel connection strategy to further reduce the search cost.

Two types of NAS methods, namely, the embedding-based NAS \cite{chatzianastasis2021graph, DBLP:conf/nips/YanZAZ020} and LSO-based NAS \cite{luo2018neural, DBLP:conf/nips/ZhangJCGC19, chatzianastasis2021graph}, both utilize the continuous embedding of neural architectures to boost the architecture optimization. The fundamental difference between the two is that whether the decoder is utilized to generate new architectures in the search phase. Representing the embedding-based method, arch2vec learns the continuous embedding of neural architectures in an unsupervised manner \cite{DBLP:conf/nips/YanZAZ020} by a GNN-based autoencoder. Afterwards, it stores the embedding of architectures in a data buffer and selects the optimal one by RL or BO strategies. However, when the strategy derives new embedding which represents a new architecture, arch2vec needs to retrieve the most similar architecture in the buffer based on the nearest neighbor algorithm, which implies that the method cannot be transferred to a large search space. 

In contrast, LSO-based NAS employs a decoder to synthesize novel architectures that might not exist during pre-training phase \cite{luo2018neural, DBLP:conf/nips/ZhangJCGC19, chatzianastasis2021graph}. While LSO-based NAS is able to derive architectures beyond the training data, potential problems may occur. First, similar embeddings may correspond to identical architectures. Second, it may yield invalid architectures that fall outside the predefined search space. This paper introduces an adaptive mechanism to mitigate these challenges.  
\subsection{Deep graph learning}
Deep graph learning (GDL) constitutes an emerging field exploring the extension of deep learning principles to data structured as graphs. Numerous studies have harnessed GDL to advance NAS. For instance, Ning et al. introduce a novel modeling approach to enhance the performance of predictor-based NAS systems \cite{ning2020generic}. Wei et al., meanwhile, advocate for the use of a graph-based predictor to augment the exploratory capabilities of evolutionary algorithm (EA)-driven NAS \cite{9723446}. 

GDL's practical significance is demonstrated in other domains. Cini et al. propose a principled probabilistic framework for learning from spatiotemporal graph data by employing score-based estimators, graph distribution models, and variance reduction methodologies \cite{cini2023sparse}. To address limitations of multi-hop aggregations, another study devises a new technique for multi-scale feature extraction on heterophilic graphs, which is achieved by developing permutation-equivariant Haar-type framelets \cite{10466590}. Prompted by the inefficiencies of batch gradient descent, the viability and utility of graph convolutional networks with random weights (GCN-RW) are investigated by adapting convolutional layers to include stochastic filters \cite{9796468}. A convolutional graph autoencoder (CGAE) has been designed for probabilistic spatio-temporal forecasting of solar irradiance \cite{10466590}. The model leverages localized spectral graph convolutions to capture graph features and variational Bayesian inference for node posterior distribution approximation. Despite both CGAE and our proposed CR-LSO/G-VAE methodologies employing graph learning in conjunction with variational autoencoder frameworks, the aim of CGAE is to apprehend the posterior distribution of nodes, and variational inference is incorporated within CGAE to approximate complex posteriors and facilitate uncertainty quantification. In contrast, we focus on neural architecture search and the distribution of latent variables to enhance the performance of gradient-based LSO. Additionally, Regan et al. outline a new framework for high-resolution remote sensing image retrieval by combining a triplet GNN with attention mechanisms and similarity-guided dictionary learning. The method optimizes both deep feature extraction and similarity assessment to elevate retrieval precision \cite{regan2023triplet}.

\section{Methodology}
This section describes the CR-LSO. The first part introduces the G-VAE which learns the continuous representations of discrete architectures. Then, the second part describes how to regularize the learning process of latent space to obtain a convex performance mapping. Finally, the gradient-based architecture optimization is presented.
\subsection{G-VAE}
As the basic work of LSO, the G-VAE transforms the search space of NAS from discrete to continuous in order to allow architecture optimization in the continuous space. The main framework of G-VAE is established based on the variational autoencoder (VAE) \cite{kingma2013auto} except that the G-VAE is designed for graph representation learning.

\subsubsection{Variational autoencoder}
Let $\alpha=(A, X, E)$ be the graph representation of an architecture $\alpha$, where $A$, $X$ and $E$ are adjacency matrix, node attributes, and edge attributes, respectively. For an architecture $\alpha$, the optimization objective of G-VAE is to learn its continuous representation $z\in \mathcal{Z}$ by learning an approximate posterior distribution $q_{\phi}(z|\alpha)$ (the encoder) and a generative model (the decoder) $p_{\theta}(\alpha|z)$, where $\phi$ and $\theta$ are learnable parameters of the encoder and decoder, respectively. This goal is achieved by Bayesian approximate inference, i.e., by minimizing the variational upper bound of the negative log-likelihood, $-{\rm log} \; p(\alpha)$, as 
\begin{align}
	\label{eq:vae_loss}
	\begin{split}
		\mathcal{L}_{\phi, \theta}(\alpha) =   \ D_{\rm KL} \big( q_{\phi}(z|\alpha) \Vert p(z)\big) \\
		+ \ \mathbb{E}_{q_{\phi}(z|\alpha)} \big[ -{\rm log} \: p_{\theta}(\alpha|z) \big],
	\end{split}
\end{align}
where $D_{\rm KL} \big( q_{\phi}(z|\alpha) \Vert p(z)\big)$ is the Kullback–Leibler (KL) divergence between the posterior $q_{\phi}(z|\alpha)$ and the isotropic Gaussian distribution $p(z)$. 

\subsection{GNN-based encoder} 
The encoder of the G-VAE is an $L$-layer graph neural network (GNN) using the synchronous message passing scheme \cite{fey2019fast}. Let $N(v)$ be the neighborhood of node $v$ in $\alpha$. Denoting $\mathbf{x}_{v}^{(k-1)}$ as the feature vector of node $v$ in layer $(k-1)$ and $e_{uv}$ as the edge feature from $u$ to $v$, the updating function of node $v$ in layer $k$ is designed as
\begin{equation}
	\label{eq:nodeupdate}
	\mathbf{x}_{v}^{(k)} = \Theta^{(k)}\mathbf{x}_{v}^{(k-1)} + {\psi}^{(k)} \big( \mathbf{x}_{v}^{(k-1)}, \mathbf{h}_{v}^{(k)} \big),
\end{equation}
where the incoming messages $\mathbf{h}_{v}^{k}$ of node $v$ is calculated by an aggregate function as
\begin{equation}
	\label{eq:aggregate}
	\mathbf{h}_{v}^{(k)} = \sum_{u\in N(v)} {\varphi}^{(k)}(\mathbf{x}_{v}^{(k-1)}, \mathbf{x}_{u}^{(k-1)}, \mathbf{e}_{uv}).
\end{equation}
In \eqref{eq:nodeupdate} and \eqref{eq:aggregate}, $\Theta^{(k)}$ is a learnable parameter set, and both ${\psi}^{(k)}(\cdot)$ and ${\varphi}^{(k)}(\cdot)$ are MLPs with nonlinear activation functions. The first term of the right hand side of \eqref{eq:nodeupdate}, $\Theta^{(k)}\mathbf{x}_{v}^{(k-1)}$, is the residual information of node update, which allows us to build a deeper GNN encoder by avoiding the vanishing gradients \cite{DBLP:conf/cvpr/HeZRS16}. Since the aggregate function (\ref{eq:aggregate}) is invariant to the permutation of incoming nodes, the encoder of G-VAE embeds an architecture into the latent space injectively. In addition, \eqref{eq:aggregate} includes an extra edge feature, which allows us to deal with the operator-on-node (e.g., the NAS-Bench-101) and operator-on-edge (e.g., the NAS-Bench-301) search spaces consistently using the same encoder. The representation of an architecture is designed as the sum of feature vectors of all nodes as 
\begin{equation}
	\textbf{h}_{\alpha} = {\rm sum}\big( \big\{  \mathbf{h}_{v}^{L} | v\in V   \big\} \big).
\end{equation}
After calculating $\mathbf{h}_{\alpha}$, two linear layers are used to output the mean $\mu(\alpha)$ and the variance $\sigma(\alpha)$ of a normal distribution, which characterize the posterior distribution $q_{\phi}(z|\alpha)$.
%\footnote{As long as the posterior is close to the prior Gaussian, it is said that the search space of NAS has been transformed into a continuous one.}

\subsection{Decoder} 
The decoder is a generative model to generate discrete architectures given latent variables $z$. However, decoding is much more difficult than encoding due to the non-Euclidean property of graphs. For methods which generate a new graph in a sequential manner, a growing graph is constructed by adding nodes and edges iteratively \cite{DBLP:conf/nips/ZhangJCGC19, you2018graphrnn}. However, these methods have to design the generation rules of discrete graphs, which is not an easy task. Inspired by \cite{simonovsky2018graphvae}, an MLP-based probabilistic graph decoder is introduced in G-VAE to generate the entire graph at once. This method is effective if the generated graphs are small, e.g., the cell-based neural architectures. 

\subsection{Convexity regularized latent space}
Note that minimizing \eqref{eq:vae_loss} does not mean that the mapping from latent space to architecture performance is convex. Thus, the latent space has to be regularized somehow to lead the underlying performance mapping to convex. In this part, the guaranteed convexity of ICNNs is leveraged to regularize the learning process of latent space. 	

\subsubsection{ICNN}
Considering a $k$-layer MLP-like neural network, the architecture of an ICNN $f_{\rm ICNN}(z)$ is defined as
\begin{equation}
	y_{i+1} = h_{i}\Big( W_{i}^{(y)}y_{i} + W_{i}^{(z)}z + b_{i} \Big), \: \ f_{\rm ICNN}(z) = y_{k},
\end{equation} 
where $i=0,...,k-1$, $y_{i}$ and $b_{i}$ are the activation output and the bias of the layer $i$, $\beta = \{ W_{0:k-1}^{(z)}, W_{0:k-1}^{(y)}, b_{0:k-1} \}$ is the learnable parameter set of ICNNs, and $h_{i}$ is nonlinear activation function. Specifically, $y_{0}$ and $W_{0}^{(y)}$ are set to zeros. Then, for such a network, the following lemma is necessary \cite{amos2017input, chen2018optimal}:

\emph{Lemma 1}: \emph{The function $f_{\rm ICNN}(z)$ is convex from input to output if all elements of $W_{1:k-1}^{y}$ are non-negative, and all activation functions $h_{i}$ are convex and non-decreasing (e.g., ReLU and LeakyReLU).}

\subsubsection{Regularizing the learning process of latent space}
To regularize the learning process of latent space, an ICNN-based predictor is used to predict the performance of neural architectures and is optimized together with the G-VAE. Then, the joint loss function of G-VAE and ICNN is 
\begin{align}
	\label{eq:vae_loss_all}
	\mathcal{L}_{\phi, \theta, \beta} & (\alpha,s) =   D_{\rm KL}\big( q_{\phi}(z|\alpha)\vert\vert  p(z) \big) + \nonumber \\ & \mathbb{E}_{q_{\phi}(z|\alpha)} \big[ -{\rm log} \ p_{\theta}(\alpha|z) + \big(s - f_{\rm ICNN}(z)\big)^{2}  \big],
\end{align}
where $s$ is the performance of architecture $\alpha$ on a particular task (e.g., the accuracy in image classification). 

Note that, the convexity of ICNN and the additional prediction loss $\big(s-f_{\rm ICNN}(z)\big)^{2}$ force the G-VAE to learn a convex performance mapping in the latent space. If the underlying performance mapping is a non-convex function, the prediction loss will not decrease to a sufficiently low level because the ICNN predictor can only approximate a convex function. In this case, in order to decrease the prediction loss, the G-VAE has to optimize the parameters of encoder and decoder to learn an appropriate embedding method which embeds the discrete architectures into a continuous convex space. 

From the above analysis, we know that the supervised training of G-VAE requires the performance of architectures. When a small number of labeled architectures are provided, it is difficult for the G-VAE to learn the continuous representations of discrete architectures.
%\footnote{Generally speaking, the training of G-VAE needs amounts of unlabeled architectures. Otherwise, the G-VAE cannot learn the effective continuous representations of architectures and cannot decode the valid discrete architectures when given the latent variables.} 
To tackle this problem, we leverage a semi-supervised approach, where a GNN predictor $g_{\rm GNN}(\alpha)$ is used to predict the architecture based on the discrete $\alpha$ to predict the performance of unlabeled architectures. The training process of G-VAE is summarized in Algorithm \ref{alg:gvae}.
\begin{algorithm}[htbp]
	\caption{Training process of G-VAE}
	{\textbf{Input:} The labeled architecture set $\mathcal{D} = \{(\alpha_{i},s_{i})\}$}, the G-VAE, the ICNN $f_{\rm ICNN}(z)$ and the GNN predictor $g_{\rm GNN}(\alpha)$. \\
	{\textbf{Begin:}}
	\begin{algorithmic}[1]
		\State{Using $\mathcal{D}$, train $g_{\rm GNN}(\alpha)$ until convergence to predict the performance of unlabeled architectures.} 
		\State{Construct a pseudo-labeled set $\mathcal{D'} = \{(\alpha_{i}',s'_{i})\}$ with $|\mathcal{D'}| \gg |\mathcal{D}|$ and $s'_{i} =  g_{\rm GNN}(\alpha_{i}')$ }. 
		\State{Construct an enlarged architecture set $\mathcal{D}_{\rm Big} = \mathcal{D} \cup \mathcal{D'}$ }. 
		\State{Using $\mathcal{D}_{\rm Big}$, train the G-VAE and $f_{\rm ICNN}(z)$ by minimizing \eqref{eq:vae_loss_all} until convergence.}
		
	\end{algorithmic}
	\textbf{Output}: The trained G-VAE and $f_{\rm ICNN}(z)$.
	\label{alg:gvae}
\end{algorithm}
\subsection{Convex neural architecture optimization}
%After the G-VAE is trained, an encoder which embeds the architectures into the latent space and a decoder which decodes latent variables to discrete architectures are obtained. What's more, the mapping from the latent space to architecture performance is regularized to be a (approximate) convex function by the regularization of the convexity of ICNN. 
\subsubsection{Architecture inference} 
The inference for preferable architectures is executed in the latent space $\mathcal{Z}$. Given an initial discrete architecture $\alpha$, we first obtain its continuous representation $z$ by the encoder. Note that $z$ is not the optimal point from the view of the predictor $f_{\rm ICNN}(z)$. Then, by moving $z$ along the gradient direction of $f_{\rm ICNN}$, we obtain a probably better architecture representation $z'$ as
\begin{equation}
	\label{eq:gradient_ascent}
	z' = z + \eta \frac{\partial f_{\rm ICNN}}{\partial z},
\end{equation}
where $\eta$ is a reasonably small step size. The process of \eqref{eq:gradient_ascent} can be performed multiple times by iteratively assigning a new $z$, thereby we obtain a set of new architecture representations $\{ z' \}$. Given each $z'$, we obtain a discrete architecture by the decoder. Note that, $\alpha'$ is better than $\alpha$, at least it is in the prediction of $f$, and this is the way for CR-LSO to continuously obtain (probably) better architectures. 

In practice, it is hard to determine the value of $\eta$. For a small $\eta$, the decoded architectures of $z'$ and $z$ maybe the same. For a large $\eta$, the decoded architecture maybe far from the (sub-)optimal one. Synthesizing these considerations, we propose an adaptive modification for $\eta$ based on whether the decoded architectures of $z$ and $z'$ are the same. To elaborate, for each optimization step, the parameter $\eta$ is initialized by $\eta \leftarrow \eta_{\rm min}$. Whenever the decoded architecture of $z'$ is in the labeled set, $\eta$ is increased by $\eta \leftarrow \eta + \eta_{\Delta}$ until the condition is satisfied or $\eta$ reaches a  maximum threshold value $\eta_{\rm max}$. However, considering the decoded architecture of $z'$ may still be an old one when $\eta$ reaches $\eta_{\rm max}$, we add Gaussian noise $\varepsilon \sim \epsilon \mathcal{N}(0, I)$ to $z$ for each optimization step to explore the search space. Analogously, the value of $\epsilon$ is modified based on an adaptive mechanism. At the beginning, $\epsilon$ is initialized by $\epsilon \leftarrow \epsilon_{\rm min}$, and whenever $\eta$ reaches $\eta_{\rm max}$ and the decoded architecture is still an old one, $\epsilon$ is enlarged by $\epsilon \leftarrow \epsilon + \epsilon_{\Delta}$. Note that we do not set a threshold value for $\epsilon$. The process for architecture inference is summarized in Algorithm \ref{algo:arch_inference}.

\begin{algorithm}[htbp]
	\caption{Architecture inference in latent space}
	\label{algo:arch_inference}
	\textbf{Input:} Labeled architecture set $\mathcal{D}$, initial step size $\eta_{\rm min}$, maximum step size $\eta_{\rm max}$, step size variation $\eta_{\Delta}$, initial noise magnitude $\epsilon_{\rm min}$, and noise magnitude variation $\epsilon_{\rm \Delta}$. \\
	\textbf{Begin:} \\
	$Not\_New \gets True$
	\begin{algorithmic}[1]
		\While{ $Not\_New$ }
			\State $z \gets z + \varepsilon$ where $ \varepsilon \sim \mathcal{N}(0, \epsilon^2)$ 
			\State  $z' \gets z + \eta \frac{\partial f_{\text{ICNN}}}{\partial z}$ 
			\Comment{gradient ascent update}
			\State $\alpha' \gets$ decode($z'$) \Comment{obtain decoded architecture}
			\If {$\alpha'$ is in labeled set $\mathcal{D}$} 
			 \State $\eta \gets \eta + \eta_{\Delta}$ \Comment{increase the step size}
			  \If {$\eta > \eta_{\text{max}}$} 
			  \State $\epsilon \gets \epsilon + \epsilon_{\Delta}$ \Comment{enlarge the noise magnitude} 
			  \EndIf
			\EndIf
		\State{$Not\_New \gets False $}
		\EndWhile
	\end{algorithmic}
	\textbf{Output}: New architecture $\alpha'$.
\end{algorithm}

%\begin{algorithm}
%	\caption{Architecture Inference in Latent Space}
%	\label{algo:arch_inference}
%	\begin{algorithmc}[1]
%		xxxx
%	\end{algorithmc}
%\end{algorithm}

%
%\footnote{In the architecture optimization phase, random noise is injected to the architecture representation $z$ to explore the search space, i.e., $z \leftarrow z + \varepsilon \mathcal{N}(0,I)$, where $\varepsilon$ is the parameter to control the variance.}

\subsubsection{Approaching the performance highland}
Although $\alpha'$ is better than $\alpha$ from the view of the predictor, the conclusion may not hold in practice, since $f_{\rm ICNN}(z)$ is a rough predictor based on the limited and noisy labeled architecture set. To obtain an accurate predictor, a natural way is to sample countless architectures from the search space and evaluate their performance on given tasks. However, it is impractical when the algorithm is executed in search space where the quantity of candidate architectures is countless. Do we really need an accurate predictor throughout the whole search space? For the purpose of architecture optimization, the answer is No, since it is not necessary to consider architectures with poor performance. Note that we are interested in high-performance architectures only, thus it is enough as long as the predictor is accurate in the performance highland. Nonetheless, it requires the NAS algorithms to continuously explore better architectures until a satisfactory architecture is obtained. However, the performance highland cannot be approached at one stroke by a static, rough prediction model. Thus, it should be completed by alternately fine-tuning the prediction model and collecting preferable candidates by step-by-step architecture inference.
	
\subsubsection{Final algorithm} Along with the above analysis, the implementation of CR-LSO is summarized in Algorithm \ref{alg:cr_lso}.  

\begin{algorithm}[htbp]
	\caption{The implementation of CR-LSO}
	{\textbf{Input:} The search space $\mathcal{I}$, the initial evaluation number $Q_{\rm start}$, the maximum evaluation number $Q_{\rm max}$}, and the size $K$ of seed set for architecture optimization. \\
	{\textbf{Begin:}}
	\begin{algorithmic}[1]
		\State{Construct a labeled architecture set $\mathcal{D}=\{ (\alpha_{i}, s_{i}) \}$, where $\alpha_{i}$ is sampled from $\mathcal{I}$ randomly, $s_{i}$ is the corresponding performance, and $|\mathcal{D}| = Q_{\rm start}$.} 
		\State{Obtain the G-VAE and $f_{\rm ICNN}(z)$ by Algorithm \ref{alg:gvae}}.
		\While{$|\mathcal{D}| < Q_{\rm max}$}
		\State {Construct an identical labeled dataset in latent space $\tilde{\mathcal{D}} = \{(z_{i},s_{i}) \}$, where $z_{i} = {\rm Encode}(\alpha_{i})$}.
		\State {Minimize the prediction loss of $f_{\rm ICNN}(z)$ on $\tilde{\mathcal{D}}$}.
		\State {Construct seed set $\mathcal{S}_{\rm seed} = \{ z_{j} \}$ with $|\mathcal{S}_{\rm seed}| = K$, where $z_{j}$ is the continuous representation of the top-$j$ architecture in $\mathcal{D}$.}
		\For{$z\in \mathcal{S_{\rm seed}}$}
		\State{Get $\alpha'$ by Algorithm \ref{algo:arch_inference}}.
		\State {Evaluate $\alpha'$ and obtain its performance $s'$}.
		\State {Append $(\alpha', s')$ to $\mathcal{D}$}.
		\EndFor
		\EndWhile
	\end{algorithmic}
	\textbf{Output}: The best architecture $\alpha^{*}$ in $\mathcal{D}$.
	\label{alg:cr_lso}
\end{algorithm}
\section{Experiments}
In this section, the effectiveness of CR-LSO is demonstrated on three NAS benchmarks, which are NAS-Bench-101 \cite{DBLP:conf/icml/YingKCR0H19}, NAS-Bench-201 \cite{DBLP:conf/iclr/Dong020}, and NAS-Bench-301 \cite{siems2020bench}. The training details of G-VAE and the hyper-parameter settings for architecture inference are included in the Appendices \ref{appendix:gvae} and \ref{appendix:ai}. 
\subsection{Architecture search in NAS-Bench-101}
\begin{table}[htbp]
	\centering
	\caption{Comparison of the state-of-the-art methods on NAS-Bench-101. }
	\begin{tabular}{lccc}
		\toprule
		Methods & Query & Test Acc. & Ranking \\
		\midrule
		RE \cite{real2019regularized} & 2000 & 93.96 $\pm$ 0.05 & 89 \\
		NAO \cite{luo2018neural} & 300 & 93.69 $\pm$ 0.06 & 1191 \\
		NAO \cite{luo2018neural} & 2000 & 93.90 $\pm$ 0.03 & 169 \\
		Yehui $et \ al.$ \cite{DBLP:conf/cvpr/TangWXCSXX0X20}  & - & 94.01 $\pm$ 0.12 & 47 \\
		SemiNAS \cite{DBLP:conf/nips/Luo0WQCL20} & 300 & 93.89 $\pm$ 0.06 & 197 \\
		SemiNAS \cite{DBLP:conf/nips/Luo0WQCL20} & 2000 & 94.02 $\pm$ 0.05 & 43 \\
		ReNAS \cite{DBLP:conf/cvpr/Xu00TJX021} & 4236 & 93.95 $\pm$ 0.11 & 93 \\ 
		\midrule
		CR-LSO (ours) & 5000 & \textbf{94.23 $\pm$ 0.00} & \textbf{2} \\ 
		CR-LSO (ours) & 2000 & 94.22 $\pm$ 1e-5 & 3 \\ 
		CR-LSO (ours) & 1000 & 94.15 $\pm$ 1e-3 & 8\\
		CR-LSO (ours) & 500 & 94.06 $\pm$ 2e-3 & 72 \\ 
		\bottomrule
	\end{tabular}
	\label{tab:nas_ben_101}
\end{table}
As the first rigorous NAS benchmark, the NAS-Bench-101 uses a cell-based search space with 423,624 architectures, each of which consists of at most 7 nodes and at most 9 edges \cite{DBLP:conf/icml/YingKCR0H19}. In fact, the NAS-Bench-101 has an operator-on-node search space, where a node denotes a specific operator and an edge denotes a connection between two distinct nodes. We represent architectures of NAS-Bench-101 using sparse graphs \cite{fey2019fast} whose nodes denote types of operators and edges denote the node connections. Since there is no edge feature for NAS-Bench-101 architectures, we set $e_{uv}=0$.  

\subsubsection{Architecture search} To test the effectiveness of CR-LSO under varying computational complexity, 4 different $Q_{\rm max}$ are used, which are 5000, 2000, 1000 and 500. The corresponding $Q_{\rm start}$ is 4000, 1500, 600 and 300, respectively. The size of seed set for architecture optimization $K=10$. All these experiments are run for 16 times independently. The architecture with the highest test accuracy among the validation top-50 architectures is reported. Table \ref{tab:nas_ben_101} shows the performance comparison in terms of query numbers. 

In terms of solution quality, CR-LSO consistently identifies architectures with superior test accuracy across various computational budgets. When allowed 5000 queries, CR-LSO attains a test accuracy of 94.23\%, which surpasses all competitors and secures the second-best ranking with minimal variance. With a significantly reduced query budget of 2000, CR-LSO manages to maintain a competitive edge, achieving an average test accuracy of 94.22\% and outperforming methods that utilized up to 2000 queries, such as RE \cite{real2019regularized} and SemiNAS \cite{DBLP:conf/nips/Luo0WQCL20}. Furthermore, CR-LSO's minimal standard deviations, especially when compared to methods like NAO \cite{luo2018neural} and ReNAS \cite{DBLP:conf/nips/Luo0WQCL20}, suggest a high level of consistency in finding high-quality solutions. This characteristic can be attributed to the stability imparted by the convexity regularization and gradient-based optimization.

%It can be seen that the proposed CR-LSO achieves the optimal NAS results. For example, when $Q_{\max} = 2000$, CR-LSO can find out the average top-3 architecture steadily. In addition, it is impressive that the standard deviation of CR-LSO is small extraordinary, which may benefit from the robustness of convexity-guaranteed gradient optimization.

\subsubsection{Latent space visualization with PCA}

\begin{figure}
	\subfloat[LS]{
		\includegraphics[width=0.22\textwidth]{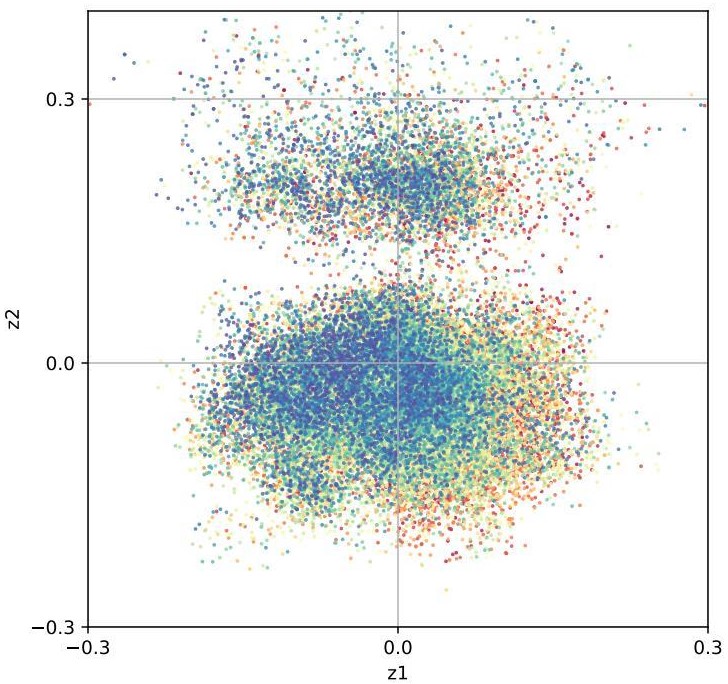}
		\label{fig:pca-a}
	}
	\subfloat[CR-LS]{
		\includegraphics[width=0.22\textwidth]{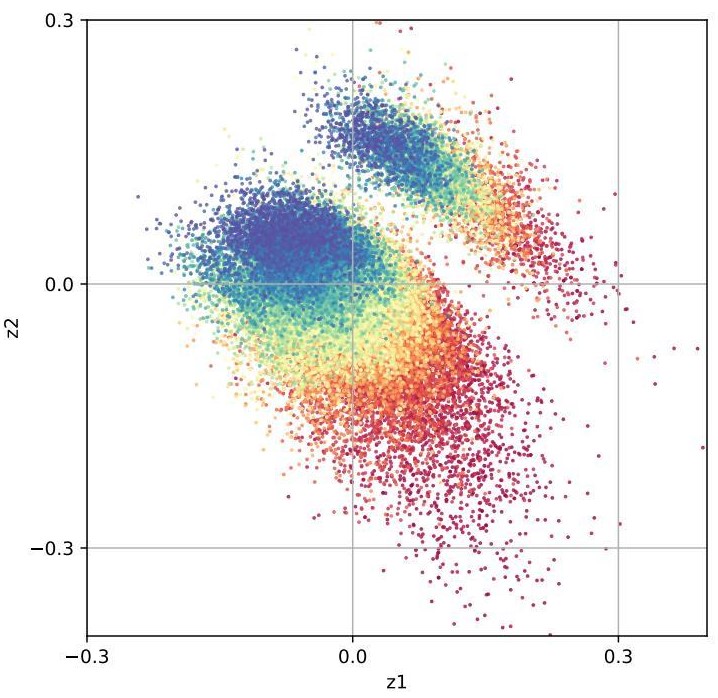}
		\label{fig:pca-b}
	}
	\caption{The PCA visualizations of architecture representations in the unconstrained latent space (LS) and the convexity regularized latent space (CR-LS) on NAS-Bench-101. The architectures with blue colors own higher rankings than those with red colors. Best  }
	\label{fig:pca}
\end{figure} 

\begin{table*}[htbp]
	\centering
	\caption{Comparison of NAS evaluation between the CR-LSO and the state-of-the-art methods on NAS-Bench-201 }
	\begin{tabular}{lcccccc}
		\toprule
		{} & \multicolumn{2}{c}{CIFAR10} & \multicolumn{2}{c}{CIFAR100} & \multicolumn{2}{c}{ImageNet16-120}   \\
		\cline{2-7} \specialrule{0em}{1.5pt}{2.0pt}
		Methods & Validation & Test & Validation & Test & Validation & Test  \\
		\midrule
		DARTS \cite{liu2018darts} & 39.77 $\pm$ 0.00 & 54.30 $\pm$ 0.00 & 15.03 $\pm$ 0.00 & 15.61 $\pm$ 0.00 & 16.43 $\pm$ 0.00 & 16.32 $\pm$ 0.00 \\
		ENAS \cite{DBLP:conf/icml/PhamGZLD18} & 37.51 $\pm$ 3.19 & 53.89 $\pm$ 0.58 & 13.37 $\pm$ 2.35 & 13.96 $\pm$ 2.33 & 15.06 $\pm$ 1.95 & 14.84 $\pm$ 2.10 \\
		GDAS \cite{DBLP:conf/cvpr/DongY19} & 89.89 $\pm$ 0.08 & 93.61 $\pm$ 0.09 & 71.34 $\pm$ 0.04 & 70.70 $\pm$ 0.30 & 41.59 $\pm$ 1.33 &  41.71 $\pm$ 0.98 \\
		RS \cite{bergstra2012random} & 90.93 $\pm$ 0.36 & 93.70 $\pm$ 0.36 & 70.93 $\pm$ 1.09 & 71.04 $\pm$ 1.07 & 44.45 $\pm$ 1.10 & 44.57 $\pm$ 1.25 \\
		REINFORCE \cite{williams1992simple} & 91.09 $\pm$ 0.37 & 93.85 $\pm$ 0.37 & 71.61 $\pm$ 1.12 & 71.71 $\pm$ 1.09 & 45.05 $\pm$ 1.02 & 45.24 $\pm$ 1.18 \\
		\midrule
		BOHB \cite{DBLP:conf/icml/FalknerKH18} & 90.82 $\pm$ 0.53 & 93.61 $\pm$ 0.52 & 72.59 $\pm$ 0.82 & 72.37 $\pm$ 0.90 & 45.44 $\pm$ 0.70 & 45.26 $\pm$ 0.83 \\
		ReNAS \cite{DBLP:conf/cvpr/Xu00TJX021} & 90.90 $\pm$ 0.31 & 93.99 $\pm$ 0.25 & 71.96 $\pm$ 0.99 & 72.12 $\pm$ 0.79 & 45.85 $\pm$ 0.47 & 45.97 $\pm$ 0.49 \\
		FairNAS \cite{DBLP:conf/iccv/Chu0X21} & 90.07 $\pm$ 0.57 & 93.23 $\pm$ 0.18 & 70.94 $\pm$ 0.94 & 71.00 $\pm$ 1.46 & 41.90 $\pm$ 1.00 & 42.19 $\pm$ 0.31 \\
		arch2vec-BO \cite{DBLP:conf/nips/YanZAZ020} & 91.41 $\pm$ 0.22 & 94.18 $\pm$ 0.24 & 73.35 $\pm$ 0.32 & 73.37 $\pm$ 0.30 & 46.34 $\pm$ 0.18 & 46.27 $\pm$ 0.37 \\
		CR-LSO (ours) & \textbf{91.54 $\pm$ 0.05} & \textbf{94.35 $\pm$ 0.05} & \textbf{73.44 $\pm$ 0.17} & \textbf{73.47 $\pm$ 0.14} & \textbf{46.51 $\pm$ 0.05} & \textbf{46.98 $\pm$ 0.35} \\
		\midrule
		Optimal & 91.61 & 94.37 & 73.49 & 73.51 & 46.77 & 47.31 \\
		\bottomrule
	\end{tabular}
	\label{tab:nas_bench_201}
\end{table*}
To illustrate the influence of convexity regularization of ICNN, we visualize the unconstrained and the convexity regularized latent spaces, respectively, in Fig. \ref{fig:pca} by projecting the architecture representations into a two-dimensional space using the PCA algorithm. It can be seen that without any constraints, the architectures with high and poor performances are highly mixed in the two-dimensional space, which means that the G-VAE has not learned a convex architecture performance mapping (Fig. \ref{fig:pca-a}). However, with the convexity regularization, the location difference of high-performance and poor-performance architectures is obvious, which means that the G-VAE has learned an approximate convex mapping in the latent space (Fig. \ref{fig:pca-b}). 
\subsection{Architecture search in NAS-Bench-201}
The NAS-Bench-201 is another popular cell-based NAS benchmark which contains 15,625 unique architectures \cite{DBLP:conf/iclr/Dong020}. Different from NAS-Bench-101, NAS-Bench-201 employs an operator-on-edge strategy where nodes denote the feature tensors of neural networks and edges denote the computational operators. Each cell consists of 4 nodes and 5 operators. The training, validation and test accuracies of each architecture are provided for three datasets including CIFAR10, CIFAR100 and ImageNet16-120. When representing an architecture of NAS-Bench-201, we denote the node attributes as the node orders of the architecture and the edge attributes as operator types. 

\subsubsection{Architecture search} We set $Q_{\rm max} = 500$ and $K=5$ to verify the effectiveness of CR-LSO. Correspondingly, we set $Q_{\rm start} = 300$ to train the GNN predictor and G-VAE. The experiments are run for 32 times independently and the evaluation results are summarized in Table \ref{tab:nas_bench_201}.

With a moderate computational budget, CR-LSO not only attains leading results across CIFAR10, CIFAR100, and ImageNet16-120 datasets, but also exhibits unparalleled consistency in its performance. On CIFAR10, CR-LSO achieves a test accuracy of 94.35\% with a minuscule standard deviation. It outperforms strong baselines like DARTS \cite{liu2018darts}, ENAS \cite{DBLP:conf/icml/PhamGZLD18} and GDAS \cite{DBLP:conf/cvpr/DongY19}, which struggle with higher variances in their results. It also surpasses optimization strategies like random search \cite{bergstra2012random} and REINFORCE \cite{williams1992simple}. Moving to CIFAR100, CR-LSO reaches a test accuracy of 73.47\% with significantly reduced variance compared to other methods. It is particularly notable than methods like BOHB \cite{DBLP:conf/icml/FalknerKH18} and FairNAS \cite{DBLP:conf/iccv/Chu0X21}, which despite showing competitive results, demonstrate higher instability in their performance. 
For ImageNet16-120, a more complex dataset, CR-LSO achieves the highest validation and test accuracies among all methods listed, with a test accuracy reaching 46.98\%. It not only beats arch2vec-BO\cite{DBLP:conf/nips/YanZAZ020}, which is a method integrating advanced embedding techniques with Bayesian optimization, but also narrows the gap to the optimal performance significantly.

%It can be seen that for all datasets, CR-LSO achieves the highest validation and test accuracies. What's more, the standard derivations of CR-LSO evaluations are significantly lower than other state-of-the-art methods. 
\subsubsection{Architecture similarity in the latent space}
\begin{figure}
	\centering
	\subfloat[Architecture similarity in LS]
	{\includegraphics[width=0.42\textwidth]{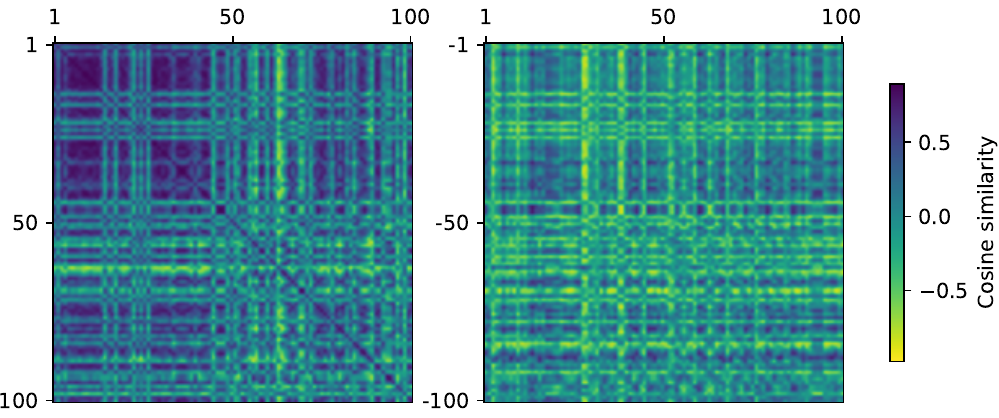}
		\label{fig:short-a}
	}
	\hfill
	\subfloat[Architecture similarity in CR-LS]
	{\includegraphics[width=0.42\textwidth]{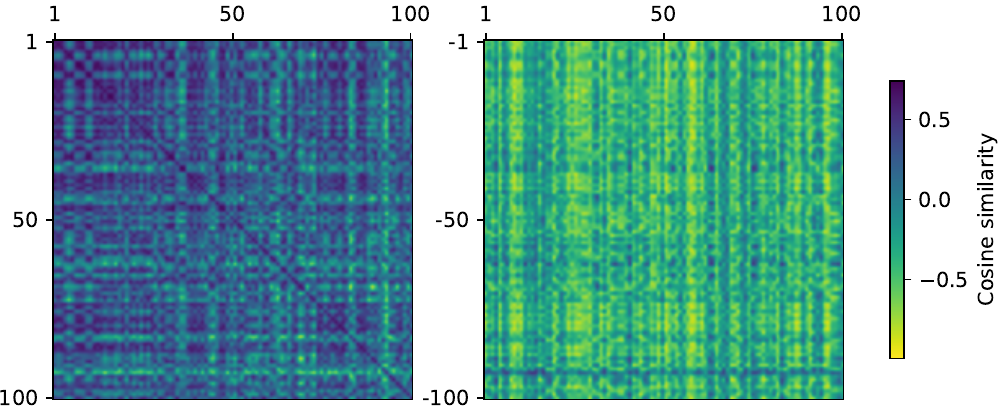}
		\label{fig:short-a}}
	\caption{The cosine similarity of architecture representations in the unconstrained latent space (LS) and the convexity regularized latent space (CR-LS) in NAS-Bench-201. Left: The architecture similarity among the optimal 100 architectures. Right: The architecture similarity between the optimal 100 and the worst 100 architectures. }
	\label{fig:short}
\end{figure} 
In this part, we test whether the architectures with similar performance are embedded into similar positions in the latent space. We pick up the optimal 100 and the worst 100 architectures from NAS-Bench-201 and visualize their cosine similarity of latent representations in Fig. \ref{fig:short}. We can see that no matter in the unconstrained or the convexity regularized latent spaces, the similarity of the optimal 100 architectures is higher than that between the top 100 and the worst 100 architectures. However, with the convexity regularization of CR-LSO, the high-performance architectures share higher similarity among themselves and share lower similarity with the worst ones. 
\subsection{Architecture search in NAS-Bench-301}
\begin{table*}[htbp]
	\centering
	\caption{Comparison of the average accuracy of top-5 architectures searched by CR-LSO and other hyper-parameter optimizations methods on NAS-Bench-301}
	\begin{tabular}{llccccc}
		\toprule
		{} & \textsc{{}} & \multicolumn{5}{c}{Query numbers of architectures}   \\
		\cline{3-7} \specialrule{0em}{1.5pt}{2.0pt}
		Search spaces & Methods &  200 & 400 & 800 & 1600 & 3200  \\
		\midrule
		Discrete & US & 94.16 $\pm$ 0.02 & 94.22 $\pm$ 0.05 & 94.29 $\pm$ 0.03 & 94.35 $\pm$ 0.04 & 94.41 $\pm$ 0.03 \\   
		\midrule
		\multirow{3}{*}{LS} & RS \cite{bergstra2012random} & 94.04 $\pm$ 0.06 & 94.19 $\pm$ 0.08 & 94.30 $\pm$ 0.05 & 94.36 $\pm$ 0.08 & 94.40 $\pm$ 0.05  \\
		{} & TPE \cite{bergstra2013making} & 94.42 $\pm$ 0.07 & 94.51 $\pm$ 0.07 & 94.63 $\pm$ 0.10 & 94.70 $\pm$ 0.10 & 94.77 $\pm$ 0.06 \\
		{} & CMA-ES \cite{nomura2021warm} & 94.34 $\pm$ 0.05 & 94.59 $\pm$ 0.15 & 94.83 $\pm$ 0.09 & 94.89 $\pm$ 0.08 & 94.92 $\pm$ 0.06 \\
		\midrule 
		\multirow{5}{*}{CR-LS} & RS \cite{bergstra2012random} & 94.17 $\pm$ 0.06 & 94.23 $\pm$ 0.05 & 94.31 $\pm$ 0.06 & 94.35 $\pm$ 0.04 & 94.42 $\pm$ 0.03 \\ 
		{} & TPE \cite{bergstra2013making} & 94.50 $\pm$ 0.06 & 94.65 $\pm$ 0.03 & 94.76 $\pm$ 0.04 & 94.81 $\pm$ 0.09 & 94.88 $\pm$ 0.06 \\ 
		{} & CMA-ES \cite{nomura2021warm} & 94.37 $\pm$ 0.09 & 94.60 $\pm$ 0.06 & 94.82 $\pm$ 0.08 & 94.93 $\pm$ 0.08 & 94.98 $\pm$  0.08 \\ 
		{} & NAO \cite{luo2018neural} & 94.49 $\pm$ 0.01 & 94.55 $\pm$ 0.04 & 94.72 $\pm$ 0.08 & 94.71 $\pm$ 0.03 & 94.75 $\pm$ 0.04 \\ 
		{} & CR-LSO (Ours) & \textbf{94.53 $\pm$ 0.04} & \textbf{94.80 $\pm$ 0.06} & \textbf{94.89 $\pm$ 0.06} & \textbf{94.94 $\pm$ 0.08} & \textbf{94.98 $\pm$ 0.02} \\ 
		\bottomrule
	\end{tabular}
	\label{tab:top-5_performance}
\end{table*}
In this section, we further verify the transfer ability of CR-LSO in a much larger NAS benchmark, i.e., the NAS-bench-301 \cite{siems2020bench}, which is a surrogate benchmark  based on the search space of DARTS \cite{liu2018darts} and contains about $10^{18}$ architectures. Similar to experiments on the NAS-Bench-201, we denote the node attributes of a graph as the node orders of the corresponding architecture and the edge attributes as the operator types. Since the normal cell and reduction cell are the same in meta-topology, we train a cell-based G-VAE to capture the continuous representation of a cell and concatenate the representations of the normal cell and the reduction cell together as the whole representation of the architecture.
\subsubsection{Baselines and experiment setup}
For fair comparison, three hyper-parameter optimization algorithms are employed as baselines including random search (RS) \cite{bergstra2012random}, BO-based tree-structured Parzen Estimator (TPE) \cite{bergstra2013making}, and EA-based covariance matrix adaptation evolution strategy (CMA-ES) \cite{nomura2021warm}. These baselines are employed under the framework of Optuna \cite{akiba2019optuna}. In addition, NAO \cite{luo2018neural} is applied as a baseline of gradient-based LSO. The uniform sampling (US) of the discrete search space is used as the lower bound of all these methods. In particular, $K=10$ and $\rho = Q_{\rm start}/Q_{\rm max} = 0.5$ are unchanged for NAO and CR-LSO. To verify whether a convexity regularized latent space (CR-LS) can promote the downstream optimization strategies, we also test the ability of RS, BO and CMA-ES in the unconstrained latent space (denoted by LS). In particular, we set $Q_{\rm max} = 200, 400, 800, 1600$ and $3200$ to test the performance of these methods under different computational complexity. The means and standard deviations of the average accuracy of the top-5 architectures are recorded. The results are calculated by five independent runs. The statistical results are summarized in Table \ref{tab:top-5_performance}.
\subsubsection{Search results analysis} 
\

\emph{\textbf{Observation 1}: Both TPE and NAO are highly competitive when $Q_{\rm max}$ is small, but CMA-ES shows much stronger competitiveness when $Q_{\rm max}$ is large.} 
When $Q_{\rm max}=200$, TPE achieves an average accuracy of $97.42\%$ and an average accuracy of $97.50\%$ in LS and CR-LS, while NAO also achieves an average accuracy of $97.49\%$ in CR-LS. However, as $Q_{\rm max}$ increases, CMA-ES significantly outperforms TPE and NAO on the architecture performance. In particular, when $Q_{\rm max}=3200$, CMA-ES obtains an average accuracy of $95.98\%$, which is higher than TPE's $94.88\%$ (in CR-LS) and $94.77\%$ (in LS) and NAO's $94.75\%$. On the other hand, RS does not perform better than US, which may because the latent space is too large for RS. 

\emph{\textbf{Observation 2}: CR-LS may promote some downstream optimization strategies.} The average accuracy of TPE in LS is higher than that in CR-LS. It means that the approximate convex mapping of CR-LS assists the TPE to find a better solution. Nevertheless, EA-based CMA-ES obtains similar performance in both LS and CR-LS, which implies that it is robust to the structure of latent space.

\emph{\textbf{Observation 3}: CR-LSO is superior to almost all listed methods in terms of sample efficiency and architecture performance.} When $Q_{\rm max} = 200$, CR-LSO achieves an average accuracy of $97.53\%$, which is slightly higher than TPE and NAO and much higher than CMA-ES. The performance of CR-LSO is improved dramatically when $Q_{\rm max} = 400$. It obtains the highest average accuracy of $97.80\%$, which significantly outperforms all listed methods. As $Q_{\rm max}$ continuously increases, CR-LSO makes no compromises on architecture performance though CMA-ES has shown strong competitiveness. 
\subsubsection{Visualization of t-SNE projection.} 
\begin{figure}[htbp]
	\centering
	\includegraphics[width=0.45\textwidth]{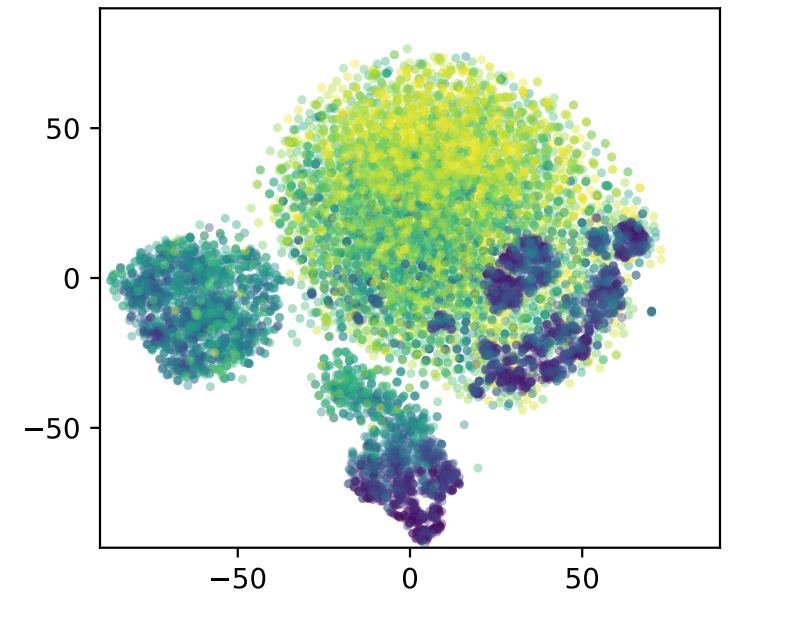}
	\includegraphics[width=0.45\textwidth]{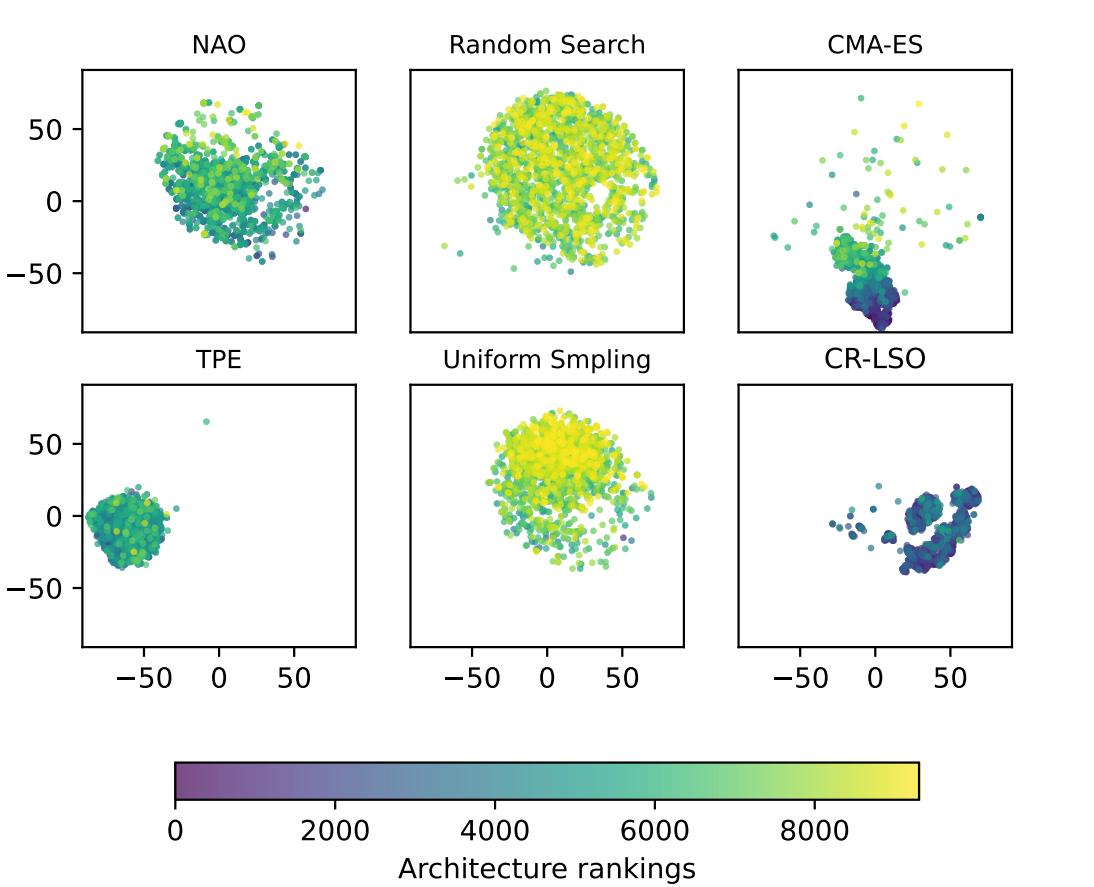}
	\caption{Two-dimensional t-SNE projection of latent representations of architectures searched by different methods. Architectures with lower rankings (darker colors) own higher performances on NAS-Bench-301.}
	\label{fig:tnse_2d}
\end{figure}
In Fig. \ref{fig:tnse_2d}, the architectures searched by different methods ($Q_{\rm max} = 3200$) are visualized by projecting the latent representations of architectures into the two-dimensional space using the t-SNE algorithm \cite{van2008visualizing}. The architectures with darker colors own better rankings. Assuming that the main characteristics of these latent representations have been preserved, we analyze the distribution difference of architectures explored by different methods. It can be seen that the architecture distributions of RS and US are quite similar in both the distribution shape and architecture performance, which matches the results in Table \ref{tab:top-5_performance}. NAO has a similar distribution shape to RS and US, but it obtains better architecture rankings because it moves the representations toward better regions along the approximated gradients. However, due to the non-convexity of conventional neural networks, it falls into the local optimal regions. Benefiting from the convexity-guaranteed gradient optimization, CR-LSO escapes from the local optimal region and explores more competitive architectures. The architectures of CR-LSO are mainly located at the boundary of the main point cloud, which might be because of the step-by-step optimization of gradients. Derivative-free methods, e.g., TPE and CMA-ES, however, share less similarity with gradient-based methods and explore architectures out of the main distribution of search space. Nevertheless, only CMA-ES explores architectures with similar competitiveness as CR-LSO.   
\begin{table}[htbp]
	\centering
	\caption{The sensitivity evaluation of $\rho = Q_{\rm start}/Q_{\rm max}$}
	\begin{tabular}{cccc}
		\toprule
		{} & \multicolumn{3}{c}{$\rho = Q_{\rm start}/Q_{\rm max}$} \\
		\cline{2-4} \specialrule{0em}{1.5pt}{2.0pt}
		$Q_{\rm max}$ & 25$\%$ & 50$\%$ & 75$\%$ \\
		\midrule
		400 & 94.70 $\pm$ 0.03 & 94.77 $\pm$ 0.06 & 94.72 $\pm$ 0.06 \\ 
		800 & 94.85 $\pm$ 0.07 & 94.86 $\pm$ 0.12 & 94.83 $\pm$ 0.09 \\ 
		1600 & 94.97 $\pm$ 0.04 & 94.95 $\pm$ 0.06 & 94.94 $\pm$ 0.05 \\
		3200 & 95.02 $\pm$ 0.03 & 95.00 $\pm$ 0.07 & 94.95 $\pm$ 0.05 \\
		\bottomrule
	\end{tabular}
	\label{tab:proportion_effect}
\end{table}
\section{Ablation Experiments}
\subsection{Hyper-parameter sensitivity evaluation}
In this part, the sensitivity of CR-LSO to some hyper-parameters is evaluated on NAS-Bench-301. The influence of the training set proportion for building the initial predictor ($\rho = Q_{\rm start}/Q_{\rm max}$) is analyzed by setting $\rho = 25\%$, $50\%$ and $75\%$. All experiments are run five times independently. Here, $K=10$ remains unchanged. From the results summarized in Table \ref{tab:proportion_effect}, we see that for $\rho =25\%$, $50\%$ and $75\%$, CR-LSO achieves similar accuracy, which means that it is robust to the change of $\rho$. 
\begin{table}[htbp]
	\centering
	\caption{The sensitivity evaluation of hyper-parameter $K$}
	\begin{tabular}{cccc}
		\toprule
		{} & \multicolumn{3}{c}{The number of $K$} \\
		\cline{2-4} \specialrule{0em}{1.5pt}{2.0pt}
		$Q_{\rm max}$ & 5 & 10 & 15 \\
		\midrule
		200 & 94.57 $\pm$ 0.05 & 94.59 $\pm$ 0.05 & 94.50 $\pm$ 0.09 \\
		400 & 94.71 $\pm$ 0.06 & 94.77 $\pm$ 0.06 & 94.73 $\pm$ 0.06 \\ 
		800 & 94.87 $\pm$ 0.02 & 94.86 $\pm$ 0.12 & 94.88 $\pm$ 0.04 \\
		1600 & 94.96 $\pm$ 0.05 & 94.95 $\pm$ 0.06 & 94.90 $\pm$ 0.09 \\
		3200 & 95.00 $\pm$ 0.05 & 95.00 $\pm$ 0.07 & 94.99 $\pm$ 0.02 \\
		\bottomrule
	\end{tabular}
	\label{tab:k_influence}
\end{table}
Next, the influence of different $K$ for architecture inference is analyzed by setting $K=5$, $10$ and $15$. Similarly, all experiments are run five times independently. Here, $\rho = 0.5$ remains unchanged. From the results summarized in Table \ref{tab:k_influence}, we see that for $K=5$, $10$ and $15$, CR-LSO achieves similar performance, meaning that it is robust to the change of $K$.

\subsection{The influence of different latent spaces}
\begin{figure}[htbp]
	\centering
	\includegraphics[width=0.46\textwidth]{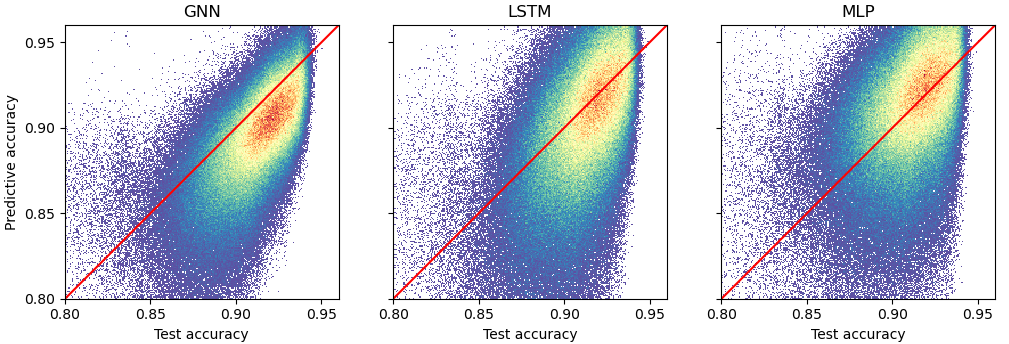}
	\caption{Visual comparison of architecture performance prediction with different VAE encoders. }
	\label{fig:pred_by_different_vaes}
\end{figure}

The natural choice for our GNN is presented in two aspects: 1) performance prediction and 2) architecture search, which are based on NAS-Bench-101. Specifically, two vanilla VAEs, whose encoders are MLP and LSTM, respectively, are employed as baselines. For MLP, the adjacency and operator matrix are flatted to vectors, which are then concatenated together as the input. For LSTM, the node connection and operator type are fed into the network alternately. 1) We train three VAEs with the same setting as in NAS-Bench-101 experiment and train MLP performance predictors in learned latent spaces. Table \ref{tab:different_vaes} and Fig. \ref{fig:pred_by_different_vaes} show the prediction performance. We can see the GNN encoder achieves the strongest pre-trained performance on accuracy prediction. 2) We execute CR-LSO based on these VAEs, and the results are summarized in Table 2. As we can see, the GNN promote the performance of CR-LSO in the case of a small query number.
\begin{table}[htbp]
	\centering
	\caption{Prediction performance in different VAE spaces on NAS-Bench-101. \% denotes the proportion of architectures for training.}
	\label{tab:prediction_performance}
	\scalebox{0.70}{\begin{tabular}{lcccccc}
			\toprule
			\specialrule{0em}{1pt}{1pt}
			\multirow{2}{*}{Models} 
			& \multicolumn{2}{c}{$0.1\%$} & \multicolumn{2}{c}{$1.0\%$} & \multicolumn{2}{c}{$10\%$} \\ 
			\cline{2-7} \specialrule{0em}{1pt}{1pt} & Spearman's $r$ & Kendall's $\tau$ & Spearman's $r$ & Kendall's $\tau$ & Spearman's $r$ & Kendall's $\tau$ \\
			\midrule \specialrule{0em}{1pt}{1pt}
			GNN & 0.67 $\pm$ 0.01 & 0.49 $\pm$ 0.01 &  0.76 $\pm$ 0.01 & 0.56 $\pm$ 0.01 & 0.87 $\pm$ 0.01 & 0.69 $\pm$ 0.01 \\
			LSTM & 0.63 $\pm$ 0.02 & 0.44 $\pm$ 0.01 & 0.72 $\pm$ 0.02 & 0.52 $\pm$ 0.01 & 0.85 $\pm$ 0.00 & 0.64 $\pm$ 0.00 \\
			MLP & 0.48 $\pm$ 0.02 & 0.32 $\pm$ 0.01 & 0.61 $\pm$ 0.01 & 0.43 $\pm$ 0.01 & 0.79 $\pm$ 0.01 & 0.58 $\pm$ 0.00 \\
			\bottomrule
	\end{tabular}}
\end{table}
\begin{table}[htbp]
	\centering
	\caption{CR-LSO performances with different VAE encoders (\%) }
	\scalebox{1.0}{
		\begin{tabular}{cccc}
			\toprule
			{} & GNN & LSTM & MLP \\
			\midrule 
			Query & Test Acc. & Test Acc. & Test Acc. \\
			500  & 94.06 $\pm$ 0.03 & 93.02 $\pm$ 0.02 & 94.94 $\pm$ 0.04 \\
			1000 & 94.15 $\pm$ 0.01 & 94.13 $\pm$ 0.02 & 94.09 $\pm$ 0.03 \\
			2000 & 94.22 $\pm$ 0.01 & 94.22 $\pm$ 0.02 & 94.14 $\pm$ 0.01 \\
			\midrule
	\end{tabular}}
	\label{tab:different_vaes}
\end{table}
\subsection{Prediction performance of ICNNs}
An accurate predictor is critical to optimize the latent variables. One may doubt whether the ICNN still has the ability to capture the underlying performance mapping. This part eliminates such doubts by analyzing the prediction performance of the ICNN quantitatively on NAS-Bench-301 \cite{siems2020bench}. To test the approximation ability, we sample 31.6K architectures from the search space of DARTS randomly and label them using the query values of NAS-Bench-301. The first 1.6K architectures are used as the training set and the other 30K ones are used as the test set. For each experiment, the training set is used to train a GNN-based predictor. Then, the predictor is used to train the G-VAE and the ICNN in a semi-supervised manner. Hereafter, the ICNN predictor is verified on the test set. Pearson and Kendall correlation coefficients are utilized to quantify the evaluation results. For comparison, the MLP is used as a baseline. If the evaluation results of the ICNN are as good as MLP's, we say that the ICNN does not lose the approximation ability. Experiments are accomplished in three different settings with the architecture number of the training set (denoted by $N$) being 100, 400 and 1600, respectively. All experiments are repeated five times independently. Means and standard deviations are recorded. The results are summarized in Table \ref{tab:prediction_performance}. As we can see, the ICNN achieves similar performance as the MLP, which indicates a solid foundation for the subsequent architecture search.
\begin{table}[htbp]
	\centering
	\caption{Comparison of prediction performance}
	\begin{tabular}{lccc}
		\toprule
		Models & $N$ & Pearson's $r$ & Kendall's $r$ \\
		\midrule
		\multirow{3}{*}{ICNN} & 100 & 0.831 $\pm$ 0.009 & 0.634 $\pm$ 0.011 \\
		{} & 400 & 0.932 $\pm$ 0.002 & 0.777 $\pm$ 0.003 \\
		{} & 1600 & 0.950 $\pm$ 0.002 & 0.812 $\pm$ 0.004 \\		
		\midrule
		\multirow{3}{*}{MLP} & 100 & 0.823 $\pm$ 0.012 & 0.629 $\pm$ 0.013 \\
		{} & 400 & 0.930 $\pm$ 0.003 & 0.773 $\pm$ 0.006 \\
		{} & 1600 & 0.948 $\pm$ 0.002 & 0.812 $\pm$ 0.004 \\
		\bottomrule
	\end{tabular}
	\label{tab:prediction_performance}
\end{table}
\subsection{The influence of ICNN regularization}
Experiments have been implemented on benchmarks 101 and 301 to show the regularization impacts of different predictors. As seen in Fig. 6, regularizing the learning process of latent spaces with an MLP predictor improves the down-stream architecture search, however, the performance improvement is not stronger than that of ICNN. 
\begin{figure}[htbp]
	\centering
	\includegraphics[width=0.38\textwidth]{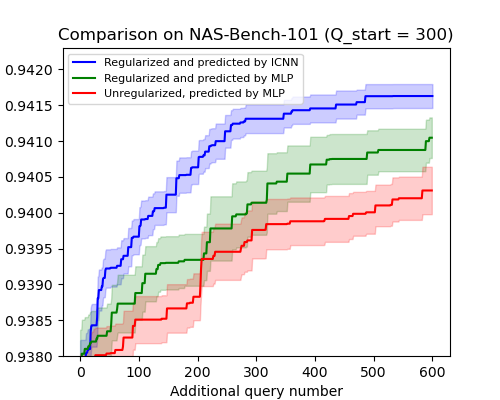}
	\includegraphics[width=0.38\textwidth]{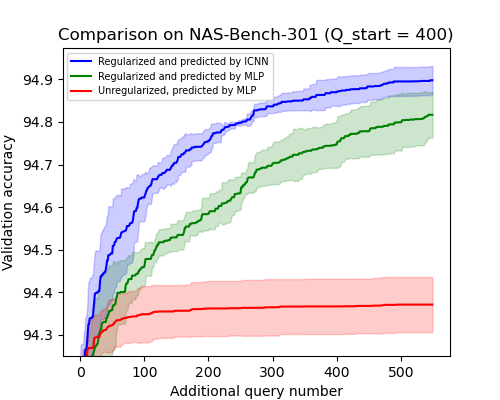}
	\caption{The comparison of regularization impacts using different predictors (best viewed in colors)}
\end{figure}
\subsection{Hyper-parameter evaluation for architecture inference}
There are numerous hyper-parameters in the phase of architecture inference, and this part is dedicated to cross-validation experiments for these hyper-parameters with a particular focus on CIFAR10 evaluations within the NAS-Bench-201 benchmark. The hyper-parameters under scrutiny include $\eta_{\rm max}$, $\eta_{\Delta}$, $\epsilon_{\rm min}$ and $\epsilon_{\rm \Delta}$. Throughout these experiments, we maintain $Q_{\rm start}=300$, $Q_{\rm max}=500$, $K=10$, and fix $\eta_{\rm min}=0.1$. It is noteworthy that the initial setting of $\eta_{\rm min}$ has minimal impact on CR-LSO's performance due to the adaptive adjustment of $\eta$ throughout the optimization process. The ranges of hyper-parameter selections are selected as: $\eta_{\rm max}\in [1.00, 5.00]$, $\eta_{\Delta} \in [0.10, 0.60]$, $\epsilon_{\rm min} \in [0.00, 0.20]$, and $\epsilon_{\Delta} \in [0.05,0.30]$. 

\begin{figure}
	\centering
	\includegraphics[width=0.8\linewidth]{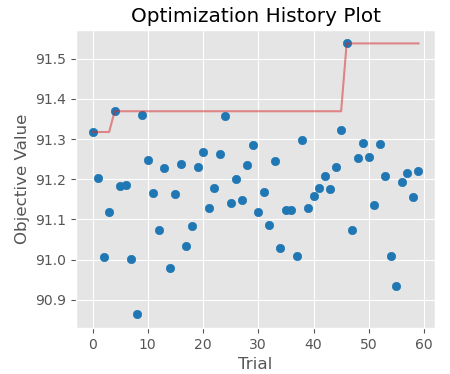}
	\caption{Optimization history for hyper-parameters on NAS-Bench-201 experiments. Along the horizontal axis, each point represents the sequential trial number in the optimization sequence. Meanwhile, the vertical axis measures the average test accuracy aggregated over 5 separate runs, which serve as the performance indicator for each set of hyper-parameters.}
	\label{fig:201history}
\end{figure}

The hyper-parameter search is implemented by Optuna \cite{akiba2019optuna} with the TPE sampler. The upper limit for the number of trials is set to 60. Notably, to circumvent potential biases in optimization, the test accuracy, as opposed to validation accuracy used in primary experiments, serves as the objective metric for this optimization. Each configuration of hyper-parameters undergoes 5 independent experiments, with the average outcome reported. The trajectory of the optimization process is depicted in Fig. \ref{fig:201history}. 

\begin{figure}
	\centering
	\includegraphics[width=1.02\linewidth]{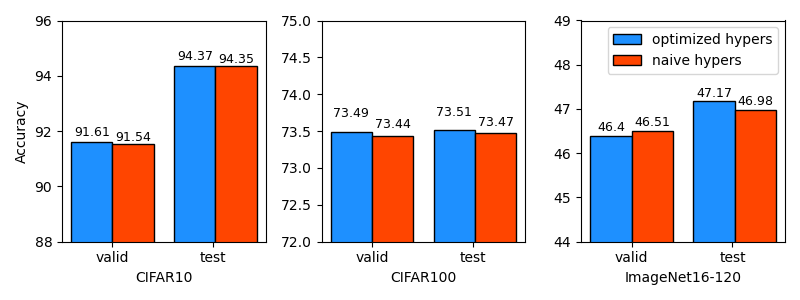}
	\caption{Performance comparison between the naive hyper-parameters and optimized hyper-parameters. The depicted results indicate a modest enhancement in performance when utilizing the optimized parameter settings compared to the naive  configurations}
	\label{fig:201}
\end{figure}

From this procedure, the identified optimal parameters are $\eta_{\rm max} = 3.16$, $\eta_{\Delta} = 0.49$, $\epsilon_{\min} = 0.02$, and $\epsilon_{\Delta} = 0.3$. These are different from those initially employed as $\eta_{\rm max} = 5.00$, $\eta_{\Delta} = 0.10$, $\epsilon_{\rm min} = 0.05$ and $\epsilon_{\Delta} = 0.05$ in the main experiments. Subsequently, we re-executed the principal experiments for both CIFAR10, CIFAR100 and ImageNet16-120 by adopting the newly optimized hyper-parameters, and the outcomes are illustrated in Fig. \ref{fig:201}. The derived hyper-parameters present more competitive results than the original ones, which implies that the performance of CR-LSO can be improved further by selecting better hyper-parameters. 
\subsection{Sensitivity evaluation for ICNN configurations}
There exist several hyper-parameters that must be specified for ICNNs, including the count of layers and hidden neurons, as well as the settings for the Adam optimizer. This segment evaluates the sensitivity of CR-LSO's performance to these hyper-parameters. Similar to previous subsection, we focus on CIFAR10 experiments in NAS-Bench-201, and use the Optuna framework to implement the selection process. Here, we denote the learning rate by $lr$, and the Adam optimizer’s momentum components by $(\beta_{1}, \beta_{2})$. Specifically, we explore the following settings. The number of layers and hidden neurons are categorical variables confined to $[2,3,4]$ and $[128,256,512]$, respectively. The learning rate $lr$ spans in a logarithmically scaled continuous interval  $[10^{-4}, 10^{-2}]$. For the momentum coefficients, $\beta_{1} $ is confined within $[0, 0.50]$ and $\beta_{2}$ ranges in $[0.50, 0.999]$. The maximum trial number is 90. 

\begin{figure}
	\centering
	\includegraphics[width=1.00\linewidth]{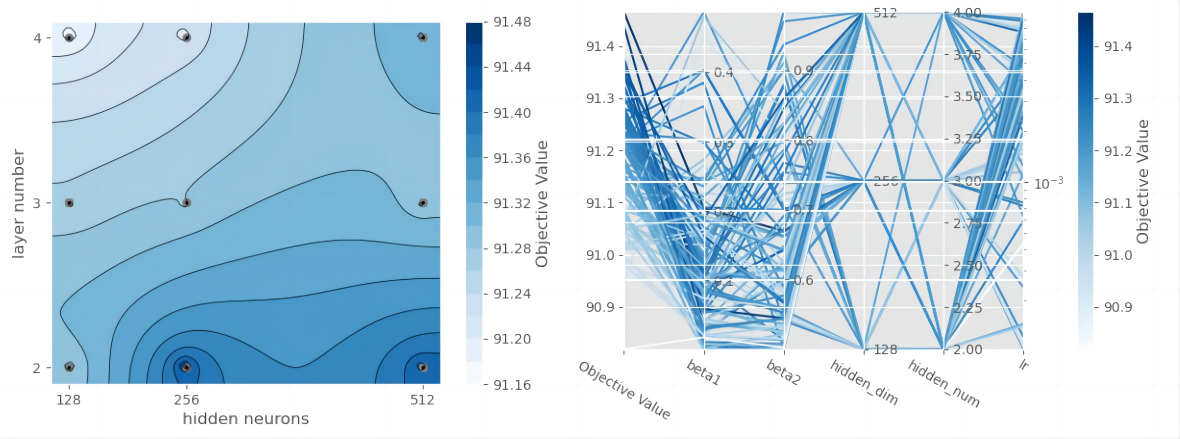}
	\caption{Sensibility evaluation for ICNN configurations and Adam optimizer. Left: comparison of ICNN configuration, which is achieved by marginalized over other variables. Right: parallel coordinate plot among all variables.}
	\label{fig:icnn}
\end{figure}

Visual depictions of the findings are presented in Fig \ref{fig:icnn}. On the left, a comparison of ICNN configurations is illustrated, which is achieved by marginalized over other variables. These results imply that ICNNs with fewer layers and more hidden neurons tend to exhibit superior performance. Conversely, configurations such as those incorporating 4 layers with 128 hidden units are less efficient. Furthermore, the parallel coordinate plotted on the right presents the parallel coordinate among all variables. It reveals preferences for $\beta_{1}$ valued below 0.3, a broad distribution of $\beta_{2}$ across its decision space, and tendency for learning rate exceeding $10^{-3}$. 
%--------------------------------

\section{Conclusion}
The paper presents an exploration of gradient-based optimization techniques applied to NAS within the latent space of a deep generative model. The contributions are as follows. 1) Considering the characteristics of neural architectures, we propose the G-VAE, a graph-based deep generative model, to learn the continuous embedding for architectures. 2)  Considering the non-convex mapping from architecture representations and their performance,  we develop the CR-LSO method, which exploits the guaranteed convexity of an ICNN to regularize the learning process of G-VAE latent space, so as to obtain an approximately convex architecture performance mapping and improve the gradient-based optimization. 

Empirical evaluations against state-of-the-art NAS methods and hyper-parameter optimization algorithms across three prevalent NAS benchmarks demonstrate the effectiveness of G-VAE and CR-LSO in terms of computational complexity and performance. It is notable that CR-LSO is able to identify the optimal architectures for CIFAR10 and CIFAR100 datasets within NAS-Bench-201 stably, which is an achievement not consistently observed by previous approaches.

\section{Prospects and Broader Impacts}
While the demonstrated performance of CR-LSO is compelling, there are aspects and future directions that deserve consideration. First, considering the computational burden of real-world NAS, it is critical to combine CR-LSO with other plugin techniques to accelerate the architecture evaluation for the real-world implementation of CR-LSO. Employing transfer learning strategies to bolster CR-LSO's scalability is promising.
Second, to tackle the challenges in generating complex graphs, developing a more robust deep generative model capable of producing neural architectures with varying nodes and connections is demanding. Third, although the inception of CR-LSO is to enhance the gradient-based optimization for NAS, it can be extended to other areas wherever the objective is to optimize a graph-based structure with some modifications. The relevant tasks may involve drug design, chip design, combinational optimizations on graphs, etc. In fact, similar methods have shown promise in enhancing neural combinatorial optimization on graphs \cite{NEURIPS2023_9c93b3cd}. These works and our CR-LSO approach suggest a broader potential for applying LSO in various domains. 

\bibliographystyle{IEEEtran}
\bibliography{mybibfile}

\begin{thebibliography}{10}
\providecommand{\url}[1]{#1}
\csname url@samestyle\endcsname
\providecommand{\newblock}{\relax}
\providecommand{\bibinfo}[2]{#2}
\providecommand{\BIBentrySTDinterwordspacing}{\spaceskip=0pt\relax}
\providecommand{\BIBentryALTinterwordstretchfactor}{4}
\providecommand{\BIBentryALTinterwordspacing}{\spaceskip=\fontdimen2\font plus
\BIBentryALTinterwordstretchfactor\fontdimen3\font minus
  \fontdimen4\font\relax}
\providecommand{\BIBforeignlanguage}[2]{{%
\expandafter\ifx\csname l@#1\endcsname\relax
\typeout{** WARNING: IEEEtran.bst: No hyphenation pattern has been}%
\typeout{** loaded for the language `#1'. Using the pattern for}%
\typeout{** the default language instead.}%
\else
\language=\csname l@#1\endcsname
\fi
#2}}
\providecommand{\BIBdecl}{\relax}
\BIBdecl

\bibitem{DBLP:conf/nips/AstudilloF21}
R.~Astudillo and P.~I. Frazier, ``Bayesian optimization of function networks,''
  in \emph{Procedings of the International Conference on 35th Neural
  Information Processing Systems}, Virtual Event, Dec. 2021, pp.
  14\,463--14\,475.

\bibitem{jumper2021highly}
J.~Jumper, R.~Evans, A.~Pritzel, T.~Green, M.~Figurnov, O.~Ronneberger,
  K.~Tunyasuvunakool, R.~Bates, A.~{\v{Z}}{\'\i}dek, A.~Potapenko
  \emph{et~al.}, ``Highly accurate protein structure prediction with
  alphafold,'' \emph{Nature}, vol. 596, no. 7873, pp. 583--589, Aug. 2021.

\bibitem{luo2018neural}
R.~Luo, F.~Tian, T.~Qin, E.~Chen, and T.~Liu, ``Neural architecture
  optimization,'' in \emph{Proceedings of the 32nd International Conference on
  Neural Information Processing Systems}, Montr{\'{e}}al, Canada, Dec. 2018,
  pp. 7827--7838.

\bibitem{DBLP:conf/nips/ZhangJCGC19}
M.~Zhang, S.~Jiang, Z.~Cui, R.~Garnett, and Y.~Chen, ``{D-VAE:} {A} variational
  autoencoder for directed acyclic graphs,'' in \emph{Proceedings of the 33rd
  International Conference on Neural Information Processing Systems},
  Vancouver, Canada, Dec. 2019, pp. 1586--1598.

\bibitem{chatzianastasis2021graph}
M.~Chatzianastasis, G.~Dasoulas, G.~Siolas, and M.~Vazirgiannis, ``Graph-based
  neural architecture search with operation embeddings,'' in \emph{Proceedings
  of the 34th {IEEE/CVF} International Conference}, Montreal, Canada, Oct.
  2021, pp. 393--402.

\bibitem{amos2017input}
B.~Amos, L.~Xu, and J.~Z. Kolter, ``Input convex neural networks,'' in
  \emph{Proceedings of the 34th International Conference on Machine Learning},
  Sydney, Australia, Aug. 2017, pp. 146--155.

\bibitem{DBLP:conf/icml/YingKCR0H19}
C.~Ying, A.~Klein, E.~Christiansen, E.~Real, K.~Murphy, and F.~Hutter,
  ``{NAS-Bench-101}: Towards reproducible neural architecture search,'' in
  \emph{Proceedings of the 36th International Conference on Machine Learning},
  Long Beach, California, {USA}, Jun. 2019, pp. 7105--7114.

\bibitem{DBLP:conf/iclr/Dong020}
X.~Dong and Y.~Yang, ``{NAS-Bench-201}: Extending the scope of reproducible
  neural architecture search,'' in \emph{Proceedings of the 8th International
  Conference on Learning Representations}, Addis Ababa, Ethiopia, Apr. 2020.

\bibitem{siems2020bench}
A.~Zela, J.~N. Siems, L.~Zimmer, J.~Lukasik, M.~Keuper, and F.~Hutter,
  ``Surrogate {NAS} benchmarks: Going beyond the limited search spaces of
  tabular {NAS} benchmarks,'' in \emph{Proceedings of the 10th International
  Conference on Learning Representations}, Virtual Event, Apr. 2022, 12 pages.

\bibitem{chen2018optimal}
Y.~Chen, Y.~Shi, and B.~Zhang, ``Optimal control via neural networks: {A}
  convex approach,'' in \emph{Proceedings of the 7th International Conference
  on Learning Representations}, New Orleans, LA, USA, May 2019, 12 pages.

\bibitem{DBLP:conf/icml/MakkuvaTOL20}
A.~V. Makkuva, A.~Taghvaei, S.~Oh, and J.~D. Lee, ``Optimal transport mapping
  via input convex neural networks,'' in \emph{Proceedings of the 37th
  International Conference on Machine Learning}, Virtual Event, Jul. 2020, pp.
  6672--6681.

\bibitem{he2021automl}
X.~He, K.~Zhao, and X.~Chu, ``Auto{ML}: A survey of the state-of-the-art,''
  \emph{Knowledge-Based Systems}, vol. 212, Jan. 2021, {A}rt. no. 106622.

\bibitem{elsken2019neural}
T.~Elsken, J.~H. Metzen, and F.~Hutter, ``Neural architecture search: A
  survey,'' \emph{The Journal of Machine Learning Research}, vol.~20, no.~1,
  pp. 1997--2017, Apr. 2019.

\bibitem{DBLP:journals/tetci/LinFCTL22}
Q.~Lin, Z.~Fang, Y.~Chen, K.~C. Tan, and Y.~Li, ``Evolutionary architectural
  search for generative adversarial networks,'' \emph{{IEEE} Transaction on
  Emerging Topic in Computational Intelligence}, vol.~6, no.~4, pp. 783--794,
  Aug. 2022.

\bibitem{10005101}
Y.~Zhou, Y.~Jin, Y.~Sun, and J.~Ding, ``Surrogate-assisted cooperative
  co-evolutionary reservoir architecture search for liquid state machines,''
  \emph{IEEE Transactions on Emerging Topics in Computational Intelligence},
  2023, early access, doi:10.1109/TETCI.2022.3228538.

\bibitem{zoph2016neural}
B.~Zoph and Q.~V. Le, ``Neural architecture search with reinforcement
  learning,'' in \emph{Proceedings of the 5th International Conference on
  Learning Representations}, Toulon, France, Apr. 2017, 12 pages.

\bibitem{DBLP:journals/ivc/JaafraLDN19}
Y.~Ja{\^{a}}fra, J.~L. Laurent, A.~Deruyver, and M.~S. Naceur, ``Reinforcement
  learning for neural architecture search: {A} review,'' \emph{Image and Vision
  Computing}, vol.~89, pp. 57--66, Sep. 2019.

\bibitem{DBLP:conf/icml/PhamGZLD18}
H.~Pham, M.~Y. Guan, B.~Zoph, Q.~V. Le, and J.~Dean, ``Efficient neural
  srchitecture search via parameter sharing,'' in \emph{Proceedings of the 35th
  International Conference on Machine Learning}, Stockholmsm{\"{a}}ssan,
  Sweden, Jul. 2018, pp. 4092--4101.

\bibitem{real2019regularized}
E.~Real, A.~Aggarwal, Y.~Huang, and Q.~V. Le, ``Regularized evolution for image
  classifier architecture search,'' in \emph{Proceedings of the 33rd {AAAI}
  Conference on Artificial Intelligence}, Honolulu, Hawaii, USA, Jan. 2019, pp.
  4780--4789.

\bibitem{DBLP:conf/eccv/NingZZWY20}
X.~Ning, Y.~Zheng, T.~Zhao, Y.~Wang, and H.~Yang, ``A generic graph-based
  neural architecture encoding scheme for predictor-based {NAS},'' in
  \emph{Proceedings of the 16th European Conference on Computer Vision},
  Glasgow, UK, Aug. 2020, pp. 189--204.

\bibitem{liu2021survey}
Y.~Liu, Y.~Sun, B.~Xue, M.~Zhang, G.~G. Yen, and K.~C. Tan, ``A survey on
  evolutionary neural architecture search,'' \emph{IEEE Transactions on Neural
  Networks and Learning Systems}, pp. 1--21, Aug. 2021.

\bibitem{zhou2019bayesnas}
H.~Zhou, M.~Yang, J.~Wang, and Pan, ``Bayes{NAS}: {A} bayesian approach for
  neural architecture search,'' in \emph{Proceedings of the 36th International
  Conference on Machine Learning}, Long Beach, California, {USA}, Jun. 2019,
  pp. 7603--7613.

\bibitem{DBLP:conf/aaai/WhiteNS21}
C.~White, W.~Neiswanger, and Y.~Savani, ``{BANANAS}: Bayesian optimization with
  neural architectures for neural architecture search,'' in \emph{Proceedings
  of the 35th {AAAI} Conference on Artificial Intelligence}, Virtual Event,
  Feb. 2021, pp. 10\,293--10\,301.

\bibitem{DBLP:conf/iclr/RuW0O21}
B.~X. Ru, X.~Wan, X.~Dong, and M.~A. Osborne, ``Interpretable neural
  architecture search via {Bayesian} optimisation with {Weisfeiler-Lehman}
  kernels,'' in \emph{Proceedings of the 9th International Conference on
  Learning Representations}, Virtual Event, May 2021, 12 pages.

\bibitem{liu2018darts}
H.~Liu, K.~Simonyan, and Y.~Yang, ``{DARTS}: Differentiable architecture
  search,'' in \emph{Proceedings of the 7th International Conference on
  Learning Representations}, New Orleans, LA, USA, May 2019, 12 pages.

\bibitem{xu2021partially}
Y.~Xu, L.~Xie, W.~Dai, X.~Zhang, X.~Chen, G.-J. Qi, H.~Xiong, and Q.~Tian,
  ``Partially-connected neural architecture search for reduced computational
  redundancy,'' \emph{IEEE Transactions on Pattern Analysis and Machine
  Intelligence}, vol.~43, no.~9, pp. 2953--2970, Sep. 2021.

\bibitem{chen2021progressive}
X.~Chen, L.~Xie, J.~Wu, and Q.~Tian, ``Progressive {DARTS}: Bridging the
  optimization gap for {NAS} in the wild,'' \emph{International Journal of
  Computer Vision}, vol. 129, no.~3, pp. 638--655, Mar. 2021.

\bibitem{DBLP:conf/bmvc/Chu021}
X.~Chu and B.~Zhang, ``Noisy differentiable architecture search,'' in
  \emph{Proceedings of the 32nd British Machine Vision Conference}, Virtual
  Event, Nov. 2021, pp. 217--231.

\bibitem{DBLP:conf/nips/YanZAZ020}
S.~Yan, Y.~Zheng, W.~Ao, X.~Zeng, and M.~Zhang, ``Does unsupervised
  architecture representation learning help neural architecture search?'' in
  \emph{Proceedings of the 34th International Conference on Neural Information
  Processing Systems}, Virtual Event, Dec. 2020, pp. 12\,486--12\,498.

\bibitem{ning2020generic}
X.~Ning, Y.~Zheng, T.~Zhao, Y.~Wang, and H.~Yang, ``A generic graph-based
  neural architecture encoding scheme for predictor-based nas,'' in
  \emph{European Conference on Computer Vision}.\hskip 1em plus 0.5em minus
  0.4em\relax Springer, 2020, pp. 189--204.

\bibitem{9723446}
C.~Wei, C.~Niu, Y.~Tang, Y.~Wang, H.~Hu, and J.~Liang, ``Npenas: Neural
  predictor guided evolution for neural architecture search,'' \emph{IEEE
  Transactions on Neural Networks and Learning Systems}, vol.~34, no.~11, pp.
  8441--8455, 2023.

\bibitem{cini2023sparse}
A.~Cini, D.~Zambon, and C.~Alippi, ``Sparse graph learning from spatiotemporal
  time series,'' \emph{Journal of Machine Learning Research}, vol.~24, no. 242,
  pp. 1--36, 2023.

\bibitem{10466590}
J.~Li, R.~Zheng, H.~Feng, M.~Li, and X.~Zhuang, ``Permutation equivariant graph
  framelets for heterophilous graph learning,'' \emph{IEEE Transactions on
  Neural Networks and Learning Systems}, pp. 1--15, 2024.

\bibitem{9796468}
C.~Huang, M.~Li, F.~Cao, H.~Fujita, Z.~Li, and X.~Wu, ``Are graph convolutional
  networks with random weights feasible?'' \emph{IEEE Transactions on Pattern
  Analysis and Machine Intelligence}, vol.~45, no.~3, pp. 2751--2768, 2023.

\bibitem{regan2023triplet}
J.~Regan and M.~Khodayar, ``A triplet graph convolutional network with
  attention and similarity-driven dictionary learning for remote sensing image
  retrieval,'' \emph{Expert Systems with Applications}, vol. 232, p. 120579,
  2023.

\bibitem{kingma2013auto}
D.~P. Kingma and M.~Welling, ``Auto-encoding variational {Bayes},'' in
  \emph{Proceedings of the 2nd International Conference on Learning
  Representations}, Banff, Canada, Apr. 2014.

\bibitem{fey2019fast}
M.~Fey and J.~E. Lenssen, ``Fast graph representation learning with {PyTorch}
  {Geometric},'' \emph{arXiv preprint, arXiv:1903.02428}, 2019.

\bibitem{DBLP:conf/cvpr/HeZRS16}
K.~He, X.~Zhang, S.~Ren, and J.~Sun, ``Deep residual learning for image
  recognition,'' in \emph{Proceedings of the 29th {IEEE} Conference on Computer
  Vision and Pattern Recognition}, Las Vegas, NV, USA, Jun. 2016, pp. 770--778.

\bibitem{you2018graphrnn}
J.~You, R.~Ying, X.~Ren, W.~L. Hamilton, and J.~Leskovec, ``Graphrnn:
  Generating realistic graphs with deep auto-regressive models,'' in
  \emph{Proceedings of the 35th International Conference on Machine Learning},
  Stockholmsm{\"{a}}ssan, Sweden, Jul. 2018, pp. 5694--5703.

\bibitem{simonovsky2018graphvae}
M.~Simonovsky and N.~Komodakis, ``{GraphVAE}: Towards generation of small
  graphs using variational autoencoders,'' in \emph{Proceedings of the 27th
  International Conference on Artificial Neural Networks}, Rhodes, Greece, Oct.
  2018, pp. 412--422.

\bibitem{DBLP:conf/cvpr/TangWXCSXX0X20}
Y.~Tang, Y.~Wang, Y.~Xu, H.~Chen, B.~Shi, C.~Xu, C.~Xu, Q.~Tian, and C.~Xu, ``A
  semi-supervised assessor of neural architectures,'' in \emph{Proceedings of
  the 33rd {IEEE/CVF} Conference on Computer Vision and Pattern Recognition},
  Seattle, USA, Jun. 2020, pp. 1807--1816.

\bibitem{DBLP:conf/nips/Luo0WQCL20}
R.~Luo, X.~Tan, R.~Wang, T.~Qin, E.~Chen, and T.~Liu, ``Semi-supervised neural
  architecture search,'' in \emph{Proceedings of the 34th International
  Conference on Neural Information Processing Systems}, Virtual Event, Dec.
  2020, pp. 10\,547--10\,557.

\bibitem{DBLP:conf/cvpr/Xu00TJX021}
Y.~Xu, Y.~Wang, K.~Han, Y.~Tang, S.~Jui, C.~Xu, and C.~Xu, ``{ReNAS}:
  Relativistic evaluation of neural architecture search,'' in \emph{Proceedings
  of the 34th {IEEE/CVF} Conference on Computer Vision and Pattern
  Recognition}, Virtual Event, Jun. 2021, pp. 4411--4420.

\bibitem{DBLP:conf/cvpr/DongY19}
X.~Dong and Y.~Yang, ``Searching for a robust neural architecture in four {GPU}
  hours,'' in \emph{Proceedings of the 32rd {IEEE/CVF} Conference on Computer
  Vision and Pattern Recognition}, Long Beach, CA, USA, Jun. 2019, pp.
  1761--1770.

\bibitem{bergstra2012random}
J.~Bergstra and Y.~Bengio, ``Random search for hyper-parameter optimization,''
  \emph{Journal of Machine Learning Research}, vol.~13, pp. 281--305, Feb.
  2012.

\bibitem{williams1992simple}
R.~J. Williams, ``Simple statistical gradient-following algorithms for
  connectionist reinforcement learning,'' \emph{Machine Learning}, vol.~8,
  no.~3, pp. 229--256, May 1992.

\bibitem{DBLP:conf/icml/FalknerKH18}
S.~Falkner, A.~Klein, and F.~Hutter, ``{BOHB}: Robust and efficient
  hyperparameter optimization at scale,'' in \emph{Proceedings of the 35th
  International Conference on Machine Learning}, Stockholmsm{\"{a}}ssan,
  Sweden, Jul. 2018, pp. 1436--1445.

\bibitem{DBLP:conf/iccv/Chu0X21}
X.~Chu, B.~Zhang, and R.~Xu, ``{FairNAS}: Rethinking evaluation fairness of
  weight sharing neural architecture search,'' in \emph{Proceedings of the 34th
  {IEEE/CVF} International Conference on Computer Vision}, Montreal, Canada,
  Oct. 2021, pp. 12\,219--12\,228.

\bibitem{bergstra2013making}
J.~Bergstra, D.~Yamins, and D.~D. Cox, ``Making a science of model search:
  Hyperparameter optimization in hundreds of dimensions for vision
  architectures,'' in \emph{Proceedings of the 30th International Conference on
  Machine Learning}, Atlanta, GA, USA, Jun. 2013, pp. 115--123.

\bibitem{nomura2021warm}
M.~Nomura, S.~Watanabe, Y.~Akimoto, Y.~Ozaki, and M.~Onishi, ``Warm starting
  {CMA-ES} for hyperparameter optimization,'' in \emph{Proceedings of the 35th
  {AAAI} Conference on Artificial Intelligence}, Virtual Event, Feb. 2021, pp.
  9188--9196.

\bibitem{akiba2019optuna}
T.~Akiba, S.~Sano, T.~Yanase, T.~Ohta, and M.~Koyama, ``Optuna: {A}
  next-generation hyperparameter optimization framework,'' in \emph{Proceedings
  of the 25th {ACM} {SIGKDD} International Conference on Knowledge Discovery
  {\&} Data Mining}, Anchorage, AK, USA, Aug. 2019, pp. 2623--2631.

\bibitem{van2008visualizing}
L.~Van~der Maaten and G.~Hinton, ``Visualizing data using {t-SNE},''
  \emph{Journal of Machine Learning Research}, vol.~9, no.~11, Nov. 2008.

\bibitem{NEURIPS2023_9c93b3cd}
Y.~Li, J.~Guo, R.~Wang, and J.~Yan, in \emph{Advances in Neural Information
  Processing Systems}, A.~Oh, T.~Naumann, A.~Globerson, K.~Saenko, M.~Hardt,
  and S.~Levine, Eds., vol.~36, 2023, pp. 50\,020--50\,040.

\end{thebibliography}

\appendix

\subsection{Training details of G-VAE} 
\label{appendix:gvae}
The architectures of NAS-Bench-101, NAS-Bench-201, and NAS-Bench-301 benchmarks are represented by sparse graphs \cite{fey2019fast}. The encoder of G-VAE is a 3-layer GNN with 512 channels (exception: 256 channels in NAS-Bench-101 due to its small scale). The dimension of latent space is 8, 64 and 128 on NAS-Bench-101, NAS-Bench-201 and NAS-Bench-301, respectively, based on the architecture complexity of different search spaces. The predictor used to regularize the learning process of latent space is a 3-layer ICNN with 256 hidden neurons. The decoder of G-VAE, which is a 3-layer MLP with 512 hidden neurons, outputs the predictive node types, edge connections, and edge types (exception: for NAS-Bench-101, there is no edge types need to be predicted) of neural architectures at once. The semi-supervised predictor is the same as the encoder but is followed by an MLP to output the predictive scalar accuracy. The semi-supervised predictor is trained for 200 epochs by an Adam optimizer with momentum coefficients $(0.0, 0.50)$, batch of size 32, and initial learning rate $10^{-4}$. The learning rate anneals to zero following a cosine scheduler. After training, we collect unlabeled architectures by sampling architectures randomly and label them using the semi-supervised predictor. For NAS-Bench-101 and NAS-Bench-201, all architectures in the search space are used to train the G-VAE. For NAS-Bench-301, 50,000 architectures are sampled from the search space and used to train the G-VAE. The G-VAE and ICNN predictor are trained for 200 epochs by an Adam optimizer with same hyper-parameters as the semi-supervised predictor except that the batch size is 512. After training, the parameters of the encoder and decoder of the G-VAE are frozen. 
\subsection{Proof of ICNN's convexity}
To prove that the output of an ICNN is convex with respect to its input, we can utilize the mathematical induction alongside the properties of convex and non-decreasing functions. The ICNN structure is described as 
\[
y_{i+1} = h_{i}\left(W_{i}^{(y)}y_{i} + W_{i}^{(z)}z + b_{i}\right), \quad f_{\rm ICNN}(z) = y_{k},
\]
where all activation functions $h_i$ are both convex and non-decreasing (like ReLU or LeakyReLU), all elements of $W_{1:k-1}^{(y)}$ are non-negative, and $y_0 = 0$ and $W_0^{(y)} = 0$.

\textbf{\emph{Base step}}: When $i=0$, we have $y_1 = h_0(W_0^{(z)}z + b_0)$. Given $h_0$ being convex and non-decreasing, $y_1$ is convex in $z$. This is because applying a non-decreasing convex function to a linear (and thus convex) function preserves convexity. 

\textbf{\emph{Inductive step}}: Assume for some $l < k$, $y_l$ is convex in $z$. We aim to show that at the next layer, $y_{l+1}$ remains convex in $z$. By definition,
\[
y_{l+1} = h_l\left(W_l^{(y)}y_l + W_l^{(z)}z + b_l\right).
\]
First, given the inductive hypothesis, $y_l$ is convex in $z$. Since $W_l^{(y)}$ has non-negative entries, we have $W_{l}^{(y)}y_l$ maintains convexity. This is because the non-negative weighted combination of convex functions remain convex. Additionally, $W_l^{(z)}z$ is linear and convex, thus adding a constant $b_l$ preserves convexity. Second, applying the convex and non-decreasing activation function $h_l$ to this convex pre-image retains convexity due to the property that the composition of a convex function with a non-decreasing convex function remains convex. Through induction, it follows that for all $i=0,1,...,k-1$, each $y_i$ is convex in $z$, which results in $f_{\rm ICNN}(z) = y_k$ also being convex in $z$. 
\subsection{Architecture inference} 
\label{appendix:ai}
Before architecture inference, the ICNN predictor is fine-tuned for 50 epochs by an Adam optimizer with the newly updated labeled set and same hyper-parameters as the semi-predictor. Other hyper-parameters are as follows: $\eta_{\rm max} = 5.00$, $\eta_{\Delta} = 0.10$, $\epsilon_{\rm min} = 0.05$ and $\epsilon_{\Delta} = 0.05$.

\end{document}